\documentclass[10pt,journal,compsoc]{IEEEtran}
\usepackage{cite}
\usepackage{amsfonts}
\usepackage{algorithmic}
\usepackage{graphicx}
\usepackage{xcolor}
\usepackage[justification=centering]{caption}
\usepackage{booktabs} 
\usepackage{subfigure}
\usepackage{amsmath}
\usepackage[normalem]{ulem}
\usepackage{enumitem}
\usepackage{multirow}
\usepackage[nomargin,inline,marginclue,draft]{fixme}
\usepackage{balance}
\usepackage{bm}
\usepackage{changepage}
\usepackage{footnote}
\usepackage{url}

\newcommand{\ie}{\emph{i.e., }}
\newcommand{\eg}{\emph{e.g., }}

\newcommand{\etc}{\emph{etc.}}
\newcommand{\wrt}{\emph{w.r.t. }}

\newlength\savedwidth

\def\BibTeX{{\rm B\kern-.05em{\sc i\kern-.025em b}\kern-.08em
    T\kern-.1667em\lower.7ex\hbox{E}\kern-.125emX}}
\begin{document}

\title{Graph Adversarial Training: Dynamically Regularizing Based on Graph Structure}

\author{Fuli~Feng,
        Xiangnan~He,
        Jie~Tang,
        Tat-Seng~Chua
\IEEEcompsocitemizethanks{
\IEEEcompsocthanksitem F. Feng and TS. Chua are with School of Computing, National University of Singapore, Computing 1, Computing Drive, 117417, Singapore.
E-mail: fulifeng93@gmail.com, dcscts@nus.edu.sg.
\protect\\
\IEEEcompsocthanksitem X. He (corresponding author) is with School of Information Science and Technology, University of Science and Technology of China, Hefei, China.
E-mail: xiangnanhe@gmail.com.
\protect\\ 
\IEEEcompsocthanksitem J. Tang is with Tsinghua University, Beijing 100084, China.
E-mail: jietang@tsinghua.edu.cn.
}
}

\markboth{IEEE TRANSACTIONS ON KNOWLEDGE AND DATA ENGINEERING, SUBMISSION 2019}
{Shell \MakeLowercase{\textit{et al.}}: Bare Demo of IEEEtran.cls for Computer Society Journals}
\IEEEtitleabstractindextext{
\begin{abstract}
Recent efforts show that neural networks are vulnerable to small but intentional perturbations on input features in visual classification tasks.
Due to the additional consideration of connections between examples (\eg articles with citation link tend to be in the same class), graph neural networks could be more sensitive to the perturbations, since the perturbations from connected examples exacerbate the impact on a target example.
\textit{Adversarial Training} (AT), a dynamic regularization technique, can resist the worst-case perturbations on input features and is a promising choice to improve model robustness and generalization. However, existing AT methods focus on standard classification, being less effective when training models on graph since it does not model the impact from connected examples.   
		
In this work, we explore adversarial training on graph, aiming to improve the robustness and generalization of models learned on graph. We propose \textit{Graph Adversarial Training} (GraphAT), which takes the impact from connected examples into account when learning to construct and resist perturbations. We give a general formulation of GraphAT, which can be seen as a dynamic regularization scheme based on the graph structure. To demonstrate the utility of GraphAT, we employ it on a state-of-the-art graph neural network model --- \textit{Graph Convolutional Network} (GCN). We conduct experiments on two citation graphs (Citeseer and Cora) and a knowledge graph (NELL), verifying the effectiveness of GraphAT which outperforms normal training on GCN by 4.51\% in node classification accuracy. Codes are available via: \url{https://github.com/fulifeng/GraphAT}.
\end{abstract}

}

\maketitle
\IEEEdisplaynontitleabstractindextext
\IEEEpeerreviewmaketitle

\section{Introduction}
\textit{Graph-based learning} makes predictions by accounting for both input features of examples and the relations between examples. It is remarkably effective for a wide range of applications, such as predicting the profiles and interests of social network users~\cite{wang2016structural,grover2016node2vec}, predicting the role of a protein in biological interaction graph~\cite{hamilton2017inductive,ying2018hierarchical}, and classifying contents like documents, videos, and webpages based on their interlinks~\cite{perozzi2014deepwalk,tang2015line,kipf2017semi}. In addition to the \textit{supervised loss} on labeled examples, graph-based learning also optimizes the \textit{smoothness} of predictions over the graph structure, that is, closely connected examples are encouraged to have similar predictions~\cite{zhu2003semi,zhou2004learning,ni2018co,velickovic2018graph}. Recently, owing to the extraordinary representation ability, deep neural networks become prevalent models for graph-based learning~\cite{yang2016revisiting,wang2016structural,kipf2017semi,ni2018co,velickovic2018graph}.

\begin{figure*}[]
	\centering
	\includegraphics[width=0.98\textwidth]{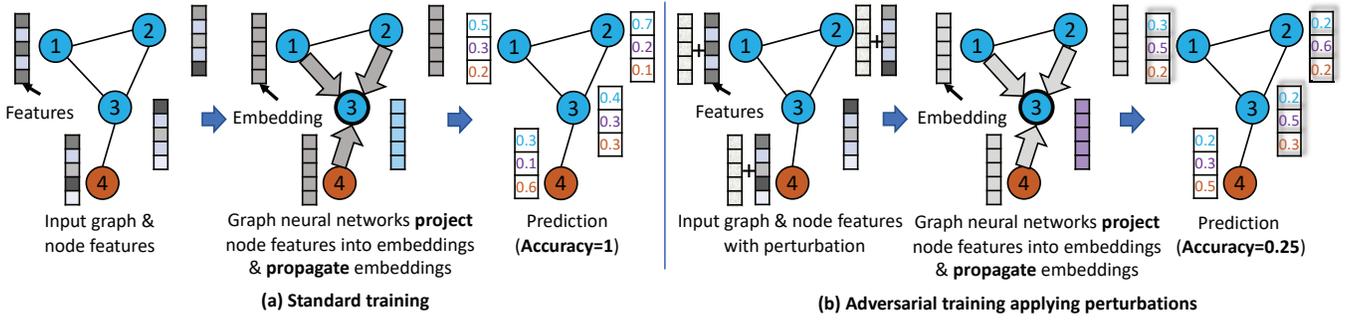} 
	\vspace{-0.2cm}
	\caption{An intuitive example to illustrate the impact of applying perturbations to the input node features on the prediction of graph neural networks. Here the model implements the graph smoothness constraint via propagating node embeddings over the graph. On the right, the model propagates the applied perturbations on the connected nodes of the target node 3, leading to a wrong prediction. Moreover, the perturbations on node 1 and 2 directly lead to the wrong associated predictions like in the standard neural networks. (Better viewed in color.)}
	\label{fig:graph_perturbation}
\end{figure*}

Despite promising performance, we argue that graph neural networks are vulnerable to small but intentional perturbations on the input features~\cite{zugner2018adversarial,Zhu2019Robust}, and this could even be more serious than the standard neural networks that do not model the graph structure. 
The reasons are twofold: 1) graph neural networks also optimize the supervised loss on labeled data, thus it will face the same vulnerability issue as the standard neural networks~\cite{goodfellow2015explaining}, and 2) the additional smoothness constraint will exacerbate the impact of perturbations, since smoothing across connected nodes\footnote{In the following sections, we interchangeably use node and example.} would aggregate the impact of perturbations from nodes connected to the target node (i.e., the node that we apply perturbations with the aim of changing its prediction). 
Figure \ref{fig:graph_perturbation} illustrates the impact of perturbations on node features with an intuitive example of a graph with 4 nodes. A graph neural network model predicts node labels (3 in total) for clean input features and features with applied perturbations, respectively. 
Here perturbations are intentionally applied to the features of nodes 1, 2, 4. Consequently, the graph neural network model is fooled to make wrong predictions on nodes 1 and 2 as with standard neural networks. Moreover, by propagating the node embeddings, the model aggregates the influence of perturbations to node 3, from which its prediction is also affected. 
In real-world applications, small perturbations like the update of node features may frequently happen, but should not change the predictions much. As such, we believe that there is a strong need to stabilize the graph neural network models during training. 

\textit{Adversarial Training} (AT) is a dynamic regularization technique that proactively simulates the perturbations during the training phase~\cite{goodfellow2015explaining}. It has been empirically shown to be able to stabilize neural networks, and enhance their robustness against perturbations in standard classification tasks~\cite{kurakin2016adversarial,miyato2016adversarial,Feng2019Enhancing,liu2019generative,Wang2019ASR,tang2019adversarial}. Therefore, employing a similar approach to that of AT on a graph neural network model would also be helpful to the model's robustness. 
However, directly employing AT on graph neural network is insufficient, since it treats examples as independent of each other and does not consider the impacts from connected examples. 
As such, we propose a new adversarial training method, named \textit{Graph Adversarial Training} (GraphAT), which learns to construct and resist perturbations by taking the graph structure into account.  

The key idea of GraphAT is that, when generating perturbations on a target example, it maximizes the divergence between the prediction of the target example and its connected examples. 
That is, the adversarial perturbations should attack the graph smoothness constraint as much as possible. 
Then, GraphAT updates model parameters by additionally minimizing a \textit{graph adversarial regularizer}, reducing the prediction divergence between the perturbed target example and its connected examples.
Through this way, GraphAT can resist the worst-case perturbations on graph-based learning and enhance model robustness. 
To efficiently calculate the adversarial perturbations, we further devise a linear approximation method based on back-propagation. 

To demonstrate GraphAT, we employ it on a well-established graph neural network model, \textit{Graph Convolutional Network} (GCN)~\cite{kipf2017semi}, which implements the smoothness constraint by performing embedding propagation. We study the method's performance on node classification, one of the most popular tasks on graph-based learning. Extensive experiments on three public benchmarks (two citation graphs and a knowledge graph) verify the strengths of GraphAT --- compared to normal training on GCN, GraphAT leads to 4.51\% accuracy improvement. Moreover, the improvements on less popular nodes (with a small degree) are more significant, highlighting the necessity of performing AT with the graph structure considered.

The main contributions of this paper are summarized as:
\begin{itemize}[leftmargin=*]
    \item We formulate \textit{Graph Adversarial Training}, a new optimization method for graph neural networks that can enhance the model's robustness against perturbations on node input features. 
    \item We devise a \textit{graph adversarial regularizer} that dynamically encourages the model to generate similar predictions on the perturbed target example and its connected examples, and develop an efficient algorithm to construct perturbations.
    \item We demonstrate the effectiveness of GraphAT on GCN, conducting experiments on three datasets which show that our method achieves state-of-the-art performance for node classification. Codes are available via: \url{https://github.com/fulifeng/GraphAT}. 
\end{itemize}

In the remainder of this paper, we first discuss related work in Section 2, followed by the problem formulation and preliminaries in Section 3. In Section 4 and 5, we elaborate the method and experimental results, respectively. We conclude the paper and envision future directions in Section 6.
\section{Related Work}
In this section, we discuss the existing research on graph-based learning and adversarial learning, which are closely related to this work.
\subsection{Graph-based Learning}
Graph, a natural representation of relational data, in which nodes and edges represent entities and their relations, is widely used in the analysis of social networks, transaction records, biological interactions, collections of interlinked documents, web pages, and multimedia contents, \etc. On such graphs, one of the most popular tasks is \textit{node classification} targeting to predicting the label of nodes in the graph by accounting for node features and the graph structure. The existing work on node classification mainly fall into two broad categories: \textit{graph Laplacian regularization} and \textit{graph embedding-based methods}. Methods lying in the former category explicitly encode the graph structure as a regularization term to smooth the predictions over the graph, \ie the regularization incurs a large penalty when similar nodes (\eg closely connected) are predicted with different labels~\cite{zhu2003semi,zhou2004learning,belkin2006manifold,talukdar2009new,feng2018learning}. 

Recently, graph embedding-based methods, which learn node embeddings that encodes the graph data, have become promising solution. Most of embedding-based methods fall into two broad categories: \textit{skip-gram based methods} and \textit{convolution based methods}, depending on how the graph data are modeled. The skip-gram based methods learn node embeddings via using the embedding of a node to predict node context that are generated by performing random walk on the graph so as the embeddings of "connected" nodes are associated to each other~\cite{perozzi2014deepwalk,tang2015line,grover2016node2vec,yang2016revisiting}. Inspired by the idea of convolution in computer vision, which aggregates contextual signals in a local window, convolution based methods iteratively aggregate representation of neighbor nodes to learn a node embedding~\cite{bruna2013spectral,duvenaud2015convolutional,defferrard2016convolutional,hamilton2017inductive,kipf2017semi,ying2018hierarchical,velickovic2018graph,ying2018graph,zhang2018deep,Wang2019NGCF}.

In both of the two categories, methods leveraging the advanced representation ability of deep neural networks (\textit{neural graph-based learning methods}) have shown remarkably effective in solving the node classification task. However, the neural graph-based learning models are vulnerable to intentionally designed perturbations indicating the unstability in generalization~\cite{dai2018adversarial,zugner2018adversarial}, and little attention has been paid to enhance the \textit{robustness} of these methods, which is the focus of this work.
\subsection{Adversarial Learning}
\subsubsection{Adversarial Training}
In order tackle the vulnerability to intentional perturbations of deep neural networks, researchers proposed adversarial training which is an alternative minimax process~\cite{szegedy2014intriguing}. The adversarial training methods augment the training process by dynamically generating adversarial examples from clean examples with perturbations maximally attacking the training objective, and then learn over these adversarial examples by minimizing an additional regularization term~\cite{dezfooli2017universal,wu2017adversarial,goodfellow2015explaining,miyato2016adversarial,miyato2018virtual,liao2018defense,tramer2018ensemble,raghunathan2019certified}. The adversarial training methods mainly fall into \textit{supervised} and \textit{semi-supervised} ones regarding the target of the training objective. In supervised learning tasks such as visual recognition~\cite{goodfellow2015explaining}, supervised loss~\cite{dezfooli2017universal,wu2017adversarial,goodfellow2015explaining} and its surrogates~\cite{liao2018defense,tramer2018ensemble,raghunathan2019certified} over adversarial examples are designed as the target of the maximization and minimization. For semi-supervised learning where partial examples are labeled, divergence of predictions for inputs around each examples is adopted as the target. Generally speaking, the philosophy of adversarial training methods is to smooth the prediction around individual inputs in a dynamical fashion.

Our work is inspired by these adversarial training methods. In addition to the local smoothness of individual examples, our method further accounts for relation between examples (\ie the graph structure) in the target of the minimax process so as to learn robust classifiers predicting smoothly over the graph structure. To the best of our knowledge, this is the first attempt to incorporate graph structure in adversarial training. 

Another emerging research topic related to our work is generating adversarial perturbations attacking neural graph-based learning models where~\cite{dai2018adversarial} and~\cite{zugner2018adversarial} are the only published work. However, methods in~\cite{dai2018adversarial} and~\cite{zugner2018adversarial} are not suitable for constructing adversarial examples in graph adversarial training. This is because these methods generate a new graph as the adversarial example for each individual node, \ie they would generate $N$ graphs when the number of nodes is $N$ leading to unaffordable memory overhead. In this work, we devise an efficient method to generate adversarial examples for graph adversarial training.
\subsubsection{Generative Adversarial Networks}
Generative adversarial networks (GAN) is a machine learning framework with two different networks as a generator and a discriminator playing minimax game on generating and detecting fake examples. Recently, several GAN-based models are proposed to learn graph embeddings, which either generate fake nodes and edges to augment embedding learning~\cite{wang2017graphgan,ding2018semi} or smooth the leaned embeddings to follow a prior distribution~\cite{sang2018aaane,yu2018learning,pan2018adversarially,dai2017adversarial}. However, using two different networks inevitably doubles the computation of model training and the labor of parameter tuning of GAN-based methods. Moreover, for different applications, one may need to build GAN from scratch, whereas our method is a generic solution can be seamlessly applied to enhance the existing graph neural network models with less computing and tuning overhead.
\section{Preliminaries}
\begin{table}[]
	\caption{Terms and notations.}
	\vspace{-0.3cm}
	\label{tab:terms}
	\resizebox{0.48\textwidth}{!}{%
		\begin{tabular}{c|l}
			\hline
			Symbol  & Definition \\ \hline \hline
			$\bm{X}=[\bm{x}_1,\bm{x}_2,\cdots,\bm{x}_N]^{T}\in\mathbb{R}^{N \times F}$     &    features of $N$ nodes.  \\ 
			$\bm{A}, \bm{D} \in \mathbb{R}^{N \times N}$ & adjacency matrix and degree matrix of a graph. \\
			$\bm{Y} = [\bm{y}_1,\bm{y}_2,\cdots,\bm{y}_M]^{T} \in \mathbb{R}^{M \times L}$ & labels of $M$ nodes. \\
			$\bm{r}_i^g$ & graph adversarial perturbation of node $i$. \\
			$\bm{r}_i^v$ & virtual graph adversarial perturbation of node $i$. \\
			$\bm{W}^l$, $\bm{b^l}$ & weights and bias to be learned at layer $l$. \\
			$\bm{\hat{y}_i} = f(\bm{x}_i, G | \mathcal{\mathbf{\Theta}})$ & prediction function. \\
			$\mathcal{\mathbf{\Theta}}$ & model parameters of the prediction function $f$.\\
			\hline
		\end{tabular}%
	}
\end{table}
We first introduce some notations used in the following sections. We use bold capital letters (e.g. $\bm{X}$) and bold lowercase letters (e.g. $\bm{x}$) to denote matrices and vectors, respectively. Note that all vectors are in a column form if not otherwise specified, and $X_{ij}$ denotes the entry of matrix $\bm{X}$ at the row $i$ and column $j$. In Table~\ref{tab:terms}, we summarize some of the terms and notations.

\subsection{Graph Representation}
The nodes and edges of a graph represent the entities of interest and their relations, respectively. First, the edges in a graph with $N$ nodes are typically represented as an adjacency matrix $\bm{A} \in \mathbb{R}^{N \times N}$. In this work, we mainly study unweighted graphs where $\bm{A}$ is a binary matrix. $A_{ij} = 1$ if there is an edge between node $i$ and $j$, otherwise $A_{ij} = 0$. Moreover, we use a diagonal matrix $\bm{D} \in \mathbb{R}^{N \times N}$ to denote the degrees of nodes, \ie $D_{ii} = \sum_{j = 1}^{N} A_{ij}$. For an attributed graph, where each node is associated with a feature vector, we use a matrix $\bm{X}=[\bm{x}_1,\bm{x}_2,\cdots,\bm{x}_N]^{T}\in\mathbb{R}^{N \times F}$ to represent the feature vectors of all nodes, where $F$ is the dimension of the features. Finally, an attributed graph is denoted as $G=(\bm{A},~\bm{D},~\bm{X})$.

\subsection{Node Classification}
On graph data, node classification is one of the most popular tasks. In the general problem setting of node classification, a graph $G$ with $N$ nodes is given, associated with labels ($\bm{Y}$) of a some portion of nodes~\cite{kipf2017semi,yang2016revisiting,velickovic2018graph}. This setting is transductive since testing nodes are observed (only features and associated edges) during training, and is the focus of this work. Here, $\bm{Y} = [\bm{y}_1,\bm{y}_2,\cdots,\bm{y}_M]^{T} \in \mathbb{R}^{M \times L}$ are the labels, where $M$ and $L$ are the numbers of labeled nodes and node classes, respectively, and $\bm{y}_i$ is the one-hot encoding of node $i$'s label. Note that, without loss of generality, we index the labeled nodes and unlabeled nodes in the range of $[1, M]$ and $(M, N]$, respectively. The target of node classification is to learn a prediction function (classifier) $\bm{\hat{y}_i} = f(\bm{x}_i, G | \mathcal{\mathbf{\Theta}})$, to forecast the label of the node, where $\mathcal{\mathbf{\Theta}}$ includes model parameters to be learned.

\subsection{Graph-based Learning}
Graph-based learning methods have been shown remarkably effective on solving the node classification task~\cite{zhu2003semi,zhou2004learning,belkin2006manifold,talukdar2009new}. Generally, most of the graph-based learning models jointly optimize two objectives: 1) \textit{supervised loss} on labeled nodes and 2) \textit{graph smoothness constraint}, which can be summarized as:
\begin{align}
	\Gamma = \Omega + \lambda \Phi,
	\label{eqn:nodeclaobj}
\end{align}
where $\Omega$ is a classification loss (\eg log loss, hinge loss, and cross-entropy loss) that measures the discrepancy between prediction and ground-truth of labeled nodes. $\Phi$ encourages \textit{smoothness} of predictions over the graph structure, which is based on the assumption that closely connected nodes tend to have similar predictions. For instance, $\Phi$ could be a \textit{graph Laplacian term}, $\sum \limits_{i,j = 1}^{N} A_{ij} \| \bm{\hat{y}}_i - \bm{\hat{y}}_j \|^2$, which directly regulates the predictions of connected nodes to be similar~\cite{zhu2003semi,zhou2004learning,belkin2006manifold,talukdar2009new}. The assumption could also be implicitly implemented by iteratively propagating \textit{node embeddings} through the graph so that connected nodes obtain close embeddings and are predicted similarly~\cite{hamilton2017inductive,kipf2017semi,ni2018co,velickovic2018graph}. Here, $\lambda$ is a hyperparameter to balance the two terms.

\section{Methodology}
In this section, we first introduce the formulation of \textit{Graph Adversarial Training}, followed by the introduction of \textit{GraphVAT}, an extension of GraphAT, which further incorporates the virtual adversarial regularization~\cite{miyato2018virtual}. We then present two solutions for the node classification task, \textit{GraphAT} and \textit{GraphVAT}, which employ GraphAT and GraphVAT to train GCN~\cite{kipf2017semi}, respectively. Finally, we analyze the time complexity of the two solutions and present the important implementation details.

\subsection{Graph Adversarial Training}
Recent advances of \textit{Adversarial Training} has been successful in learning deep neural network-based classifiers, making them robust against perturbations for a wide range of standard classification tasks such as visual recognition~\cite{goodfellow2015explaining,kurakin2016adversarial,miyato2018virtual} and text classification~\cite{miyato2016adversarial}. Generally, applying AT would regulate the model parameters to smooth the output distribution~\cite{miyato2018virtual}. Specifically, for each clean example in the dataset, AT encourages the model to assign similar outputs to the artificial input (\ie the \textit{adversarial example}) derived from the clean example. Inspired by the philosophy of standard AT, we develop graph adversarial training, which trains graph neural network modules in the manner of generating adversarial examples and optimizing additional regularization terms over the adversarial examples, so as to prevent the adverse effects of perturbations. Here the focus is to prevent perturbations propagated through node connections (as illustrated in Figure~\ref{fig:graph_perturbation}), \ie accounting for graph structure in adversarial training.

Generally, the formulation of GraphAT is:
\begin{align}
\small
& \text{\textbf{min:}}~\Gamma_{GAT} = \Gamma + \beta \sum_{i=1}^N \sum_{j \in \mathcal{N}_i} d(f(\bm{x}_i + \bm{r}_i^{g}, G | \mathcal{\mathbf{\Theta}}), f(\bm{x}_j, G | \mathcal{\mathbf{\Theta}})), \notag \\
& \text{\textbf{max:}}~\bm{r}_i^{g} = \arg\max_{\bm{r}_i, \| \bm{r}_i \| \leq \epsilon} \sum_{j \in \mathcal{N}_i} d(f(\bm{x}_i + \bm{r}_i, G | \mathcal{\mathbf{\hat{\Theta}}}), f(\bm{x}_j, G | \mathcal{\mathbf{\hat{\Theta}}})), 
\label{eqn:objgat}
\end{align}
where $\Gamma_{GAT}$ is the training objective function with two terms: the standard objective function of the origin graph-based learning model (\eg Equation~\ref{eqn:nodeclaobj}) and \textit{graph adversarial regularizer}. The second term encourages the graph adversarial examples to be classified similarly as connected examples where $\mathcal{\mathbf{\Theta}}$ denotes the parameters to be learned, and $d$ is a nonnegative function that measures the divergence (\eg Kullback-Leibler divergence~\cite{Joyce2013Kullback}) between two predictions. $\bm{r}_i^{g}$ denotes the graph adversarial perturbation, which is applied to the input feature of the clean example $i$ to construct a graph adversarial example.

The graph adversarial perturbation is calculated by maximizing the graph adversarial regularizer under current value of model parameters. That is to say, the graph adversarial perturbation is the direction of changes on the input feature, which can maximally attack the graph adversarial regularizer, \ie the worst case of perturbations propagated from neighbor nodes. $\epsilon$ is a hyperparameter controling the magnitude of perturbations, which is typically set as small values so that the feature distribution of adversarial examples is close to that of clean examples.

Generally, similar to the standard AT, each iteration of GraphAT can also be viewed as playing a \textit{minimax game}:
\begin{itemize}[leftmargin=*]
	\item \textbf{Maximization}: GraphAT generates graph adversarial perturbations from clean examples, which break the smoothness between connected nodes to the maximum extent. and then constructs graph adversarial examples by adding the perturbations to the input of associated clean examples.
	\item \textbf{Minimization}: GraphAT minimizes the objective function of the graph neural network with an additional regularizer over graph adversarial examples, by encouraging smoothness between predictions of adversarial examples and connected examples. As such, the model becomes robust against perturbations propagated through the graph.
\end{itemize}
\begin{figure}[]
	\centering
	\includegraphics[width=0.48\textwidth]{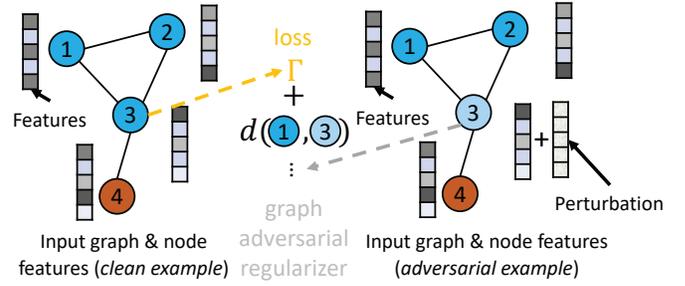} 
	\vspace{-0.2cm}
	\caption{The training process of GraphAT: 1) constructing graph adversarial example and 2) updating model parameters by minimizing loss and graph adversarial regularizer.}
	\label{fig:graph_train}
\end{figure}
Figure~\ref{fig:graph_train} illustrates the process of GraphAT. While the traditional graph-based regularizations (\eg the graph Laplacian term) also encourage the smoothness of predictions over the graph structure, GraphAT is believed to be a more advanced regulation for two reasons: 1) the regularization performed by GraphAT is dynamic since the adversarial examples are adaptively generated according to the current parameters and predictions of the model whereas the standard graph-based regularizations are static; and 2) GraphAT to some extent augments the training data, since the generated adversarial examples have not occurred in the training data, which is beneficial to model generalization.


\textbf{Approximation.} It is non-trivial to obtain the closed-form solution of $\bm{r}_i^{g}$. Inspired by the \textit{linear approximation} method proposed in~\cite{goodfellow2015explaining} for standard adversarial training, we also design a linear approximation method to calculate the graph adversarial perturbations in GraphAT, of which the formulation is:
\begin{align}
\bm{r}_i^{g} \approx \epsilon \frac{\bm{g}}{\| \bm{g} \|} \text{, where } \bm{g} = \nabla_{\bm{x}_i} \sum_{j \in \mathcal{N}_i} D(f(\bm{x}_i, G | \bm{\hat{\Theta}}), f(\bm{x}_j, G | \bm{\hat{\Theta}})),
\label{eqn:solgat}
\end{align}
where $\bm{g}$ is the gradient \wrt the input $\bm{x}_i$. For graph neural network models, the gradient can be efficiently calculated by one backpropagation. Note that $\bm{\hat{\Theta}}$ is a constant set denoting the current model parameters.

\subsection{Virtual Graph Adversarial Training}
Considering that node classification is a task of semi-supervised learning by nature, we further devise an extended version of GraphAT, named GraphVAT which additionally smooths the distribution of predictions around each clean example to further enhance the model robustness. Inspired by the idea of virtual adversarial training~\cite{miyato2018virtual}, we further add a virtual adversarial regularizer into the training objective function and construct virtual adversarial examples to attack the local smoothness of predictions. Compared to standard AT which only considers labeled clean examples, virtual adversarial training additionally encourages the model to make consistent predictions around the unlabeled clean examples. The formulation of GraphVAT is:
\begin{align}
& \text{\textbf{min:}}~\Gamma_{GVAT} = \Gamma + \underbrace{\alpha \sum_{i=1}^N d(f(\bm{x}_i + \bm{r}_i^{v}, G | \bm{\Theta}), \bm{\tilde{y}}_i)}_{\text{virtual adversarial regularizer}} + \notag \\
& \quad \quad \quad \quad \quad \underbrace{\beta \sum_{i=1}^N \sum_{j \in \mathcal{N}_i} d(f(\bm{x}_i + \bm{r}_i^{g}, G | \bm{\Theta}), f(\bm{x}_j, G | \bm{\Theta}))}_{\text{graph adversarial regularizer}}, \notag \\
& \text{\textbf{max:}}~\bm{r}_i^{v} = \arg\max_{\bm{r}_i', \| \bm{r}_i' \| \leq \epsilon'} d(f(\bm{x}_i + \bm{r}_i', G | \bm{\hat{\Theta}}), \bm{\tilde{y}}_i), 
\label{eqn:objgatv}
\end{align}
where $\bm{r}_i'$ denotes the virtual adversarial perturbation, the direction that leads to the largest change on the model prediction of $\bm{x}_i$. For labeled nodes and unlabeled nodes, $\bm{\tilde{y}}_i$ denotes ground truth label and model prediction, respectively. That is,
$$ \bm{\tilde{y}}_i = \left\{
\begin{aligned}
&\bm{\hat{y}}_i,~&i \leq M~\text{(labeled node)}, \\
&f(\bm{x}_i, G | \bm{\hat{\Theta}}),~&M < i \leq N~\text{(unlabeled node)}. \\
\end{aligned}
\right.
$$
Note that GraphVAT can be seen as jointly playing two minimax games with three players, where the two maximum players generate virtual adversarial examples and graph adversarial examples, respectively. That is, in each iteration, two types of perturbations and the associated adversarial examples are generated to attack 1) the smoothness of prediction around individual clean example; and 2) the smoothness of connected examples, respectively. By minimizing the additional regularizers over these adversarial examples, the learned model is encouraged to be more smooth and robust, achieving good generalization.

\textbf{Approximation.} For labeled nodes, $\bm{r}_i'$ can be easily evaluated via linear approximation~\cite{goodfellow2015explaining}, \ie calculating the gradient of $d(f(\bm{x}_i, G | \bm{\hat{\Theta}}), \bm{\tilde{y}}_i)$ \wrt $\bm{x}_i$. For unlabeled nodes, such approximation is infeasible since the gradient will always be zero. This is because $d(f(\bm{x}_i, G | \bm{\hat{\Theta}}), \bm{\tilde{y}}_i)$ achieves the minimum value (0) at $\bm{x}_i$ (note that $\bm{\tilde{y}}_i = f(\bm{x}_i, G | \bm{\hat{\Theta}})$ for unlabeled data). Realizing that the first-order gradient is always zero, we estimate $\bm{r}_i'$ from the second-order Taylor approximation of $d(f(\bm{x}_i + \bm{r}_i', G | \bm{\hat{\Theta}}), \bm{\tilde{y}}_i)$. That is, $\bm{r}_i^{v} \approx \arg\max_{\bm{r}_i', \| \bm{r}_i' \| \leq \epsilon'} \frac{1}{2} \bm{r}_i'^T \bm{H} \bm{r}_i'$ where $\bm{H}$ is the Hessian matrix of $d(f(\bm{x}_i + \bm{r}_i', G | \bm{\hat{\Theta}}), \bm{\tilde{y}}_i)$. For the consideration of efficiency, we calculate $\bm{r}_i^{v}$ via the power iteration approximation~\cite{miyato2018virtual}:
\begin{align}
\bm{r}_i^{v} \approx \epsilon' \frac{\bm{g}}{\| \bm{g} \|} \text{, where } \bm{g} = \nabla_{\bm{r}_i} d(f(\bm{x}_i + \bm{r}_i, G | \bm{\hat{\Theta}}, \bm{\tilde{y}}_i)) \left |_{\bm{r}_i = \xi \bm{d}}\right.,
\end{align}
where $\bm{d}$ is a random vector. 
Detailed derivation of the method is referred to~\cite{miyato2018virtual}.

\subsection{Graph Convolution Network}
Inspired by the extraordinary representation ability, many neural networks have been used as the predictive model $f(\bm{x}_i, G | \bm{\Theta})$~\cite{wang2016structural,kipf2017semi,ni2018co,velickovic2018graph}. Under the transductive setting, Graph Convolutional Network~\cite{kipf2017semi} is a state-of-the-art model. Specifically, GCN stacks multiple graph convolution layers, which is formulated:
\begin{align}
\bm{H}^{l} = \sigma \left ( \bm{\widetilde{D}}^{-\frac{1}{2}} \bm{\widetilde{A}} \bm{\widetilde{D}}^{-\frac{1}{2}} \left ( \bm{H}^{l - 1} \bm{W}^{l} + \bm{b}^{l} \right ) \right ).
\end{align}
Specifically, the $l$-th graph convolution layer conducts three operations to project $\bm{H}^{l - 1} \in \mathbb{R}^{N \times D^{l - 1}}$ (the output of the $(l - 1)$-th layer or the node features $\bm{X}$) into $\bm{H}^{l} \in \mathbb{R}^{N \times D^{l}}$, where $D^{l - 1}$ and $D^{l}$ are the output dimension of layer $l - 1$ and $l$, respectively. 
\begin{itemize}[leftmargin=*]
	\item Similar as the fully connected layer, the graph convolution layer first \textit{projects} the input ($\bm{H}^{l - 1}$) into latent representations with $\bm{W}^{l} \in \mathbb{R}^{D^{l - 1} \times D 
		^{l}}$ and $\bm{b}^{l} \in \mathbb{R}^{D^{l}}$. 
	\item It then \textit{propagates} the latent representations ($\bm{H}^{l - 1} \bm{W}^{l} + \bm{b}^{l}$) through the normalizied adjacency matrix $\bm{\widetilde{D}}^{-\frac{1}{2}} \bm{\widetilde{A}} \bm{\widetilde{D}}^{-\frac{1}{2}}$ with self-connections, where $\bm{\widetilde{D}} = \bm{D} + \bm{I}$ and $\bm{\widetilde{A}} = \bm{A} + \bm{I}$ ($\bm{I} \in \mathbb{R}^{N \times N}$ is an identity matrix). Here, the representation of node $i$ in $\bm{H}$ is the aggregation of latent representations in ($\bm{H}^{l - 1} \bm{W}^{l} + \bm{b}^{l}$) of nodes connected to $i$ (including itself due to the self-connection). 
	\item Finally, a non-linear activation function $\sigma$ (\eg the sigmoid, hyperbolic tangent, and rectifier functions) is applied to allow non-linearity.
\end{itemize}

The original objective function of GCN is,
\begin{align}
\sum_{i=1}^{M} cross\text{-}entropy(f(\bm{x}_i, G | \bm{\Theta}), \bm{y}_i) + \lambda \| \bm{\Theta} \|_F^2,
\label{eqn:objgcn}
\end{align}
where the second term is $L_2$-norm to prevent overfitting. 
To train GCN with our proposed GraphAT and GraphVAT, we set the $\Gamma$ term in Equation~\ref{eqn:objgat} and~\ref{eqn:objgatv} as the cross-entropy loss in Equation~\ref{eqn:objgcn}, which are minimized to update the parameter of GCN, respectively. 

\subsection{Time Complexity and Implementation}
\label{ss:time_comp}
\textit{Time Complexity.} As compared to GCN with standard training, the additional computation of GraphAT is twofold: 1) generating graph adversarial perturbations ($\{\bm{r}_i^g, i < N\}$) with Equation~\ref{eqn:solgat} and 2) calculating the value of graph adversarial regularizer ($\sum_{i=1}^N \sum_{j \in \mathcal{N}_i} D(f(\bm{x}_i + \bm{r}_i^{g}, G | \bm{\Theta}), f(\bm{x}_j, G | \bm{\Theta}))$). Considering that they can be accomplished with a back-propagation (to calculate $\bm{r}_i^{g}$) and a forward-propagation (to calculate $f(\bm{x}_i + \bm{r}_i^{g}, G | \bm{\Theta})$), the computation overhead of GraphAT is acceptable. Additionally, GraphVAT computes virtual adversarial perturbations and virtual adversarial regularizer, which can also be finished by performing 
one back-propagation and one forward-propagation~\cite{miyato2018virtual}. It indicates that the overhead of GraphVAT is still acceptable~\cite{miyato2018virtual}.

\textit{Implementation.} Noting that number of connected nodes varies a lot across the nodes in the graph, we sample $K$ neighbors for each node to generate adversarial examples and calculate the graph adversarial regularizer to facilitate the calculation.
Here, the following sampling strategies~\cite{ying2018graph} are considered:
\begin{itemize}[leftmargin=*]
	\item \textbf{Uniform:} neighbors are selected uniformly.
	\item \textbf{Degree:} the probability of selecting a node is proportional to the normalized node degree.
	\item \textbf{Degree-Reverse:} on the contrary, the probability is the reciprocal of node degree (also normalized to sum to unity).
	\item \textbf{PageRank:} it performs PageRank~\cite{page1999pagerank} on the graph and takes the normalized pagerank score as the sampling probability.
\end{itemize}
Note that advanced but complex sampling strategies (\eg the one proposed in~\cite{ying2018graph}) are not considered due to efficiency consideration.

\section{Experiments}
\subsection{Experimental Settings}
\subsubsection{Datasets}
We follow the same experimental settings as in~\cite{kipf2017semi} and conduct experiments on two types of node classification datasets: citation network datasets (Citeseer and Cora~\cite{sen2008collective}) and knowledge graph (NELL~\cite{yang2016revisiting})\footnote{\url{https://github.com/kimiyoung/planetoid}.}, of which the statistics are summarized in Table~\ref{tab:dataset}.

\begin{itemize}[leftmargin=*]
    \item In the citation networks, nodes and edges represent documents and citation links between documents, respectively. Note that the direction of edge is omitted since a citation is assumed to have equally impacts on the prediction of the associated two documents. Each document is associated with a normalized bag-of-words feature vector and a class label. During training, we use features of all nodes, but only 20 labels per class. 500 and 1,000 of the remaining nodes are used as validation and testing, respectively.
    \item NELL is a bipartite graph of 55,864 relation nodes and 9,891 entity nodes, extracted from a knowledge graph which is a set of triplets in the format of $(e_1, r, e_2)$. Here $e_1$ and $e_2$ are entities, and $r$ is the connected relation between them. Following~\cite{kipf2017semi}, each relation $r$ is split into two relation nodes ($r_1$ and $r_2$), from which two edges $(e_1, r_1)$ and $(e_2, r_2)$ are constructed. Entity nodes and relation nodes are described by bag-of-words feature vectors (normalized) and one-hot encodings, respectively. Note that we pad zero values to align the feature vectors of entity and relation nodes. 
    Here only labels of entity nodes are available and only $0.001$ of entities under each class are labeled during training.
\end{itemize}
\begin{table}[]
\caption{Statistics of the experiment datasets.}
\vspace{-0.3cm}
\label{tab:dataset}
\resizebox{0.48\textwidth}{!}{%
\begin{tabular}{c|ccccc}
\hline
Dataset  & \#Nodes & \#Edges & \#Classes & \#Features & Label rate \\ \hline \hline
Citeseer & 3,312   & 4,732   & 6         & 3,703      & 0.036      \\ 
Cora     & 2,708   & 5,429   & 7         & 1,433      & 0.052      \\ 
NELL     & 65,755  & 266,144 & 210       & 5,414      & 0.001      \\ \hline
\end{tabular}%
}
\end{table}

\subsubsection{Baselines}
We compare the following baselines:
\begin{itemize}[leftmargin=*]
	\item \textbf{LP}~\cite{zhu2003semi}: label propagation ignores node features and only propagates labels over the graph structure.
	\item \textbf{DeepWalk}~\cite{perozzi2014deepwalk}: it is a skip-gram based graph embedding method, which learns the embedding of a node by predicting its contexts that are generated by performing random walk on the graph.
	\item \textbf{SemiEmb}~\cite{weston2012deep}: it learns node embeddings from node features and leverages Laplacian regularization to encourage connected nodes have close embeddings.
	\item \textbf{Planetoid}~\cite{yang2016revisiting}: similar as DeepWalk, this method learns node embeddings by predicting node context, but additionally accounts for node features.
	\item \textbf{GCN}~\cite{kipf2017semi}: it stacks two graph convolution layers to project node features into labels. It propagates node representations and predictions over the graph structure to smooth the output.
	\item \textbf{GraphSGAN}~\cite{ding2018semi}: this is a semi-supervised generative adversarial network which encodes the density signal of the graph structure during generation of fake nodes.
\end{itemize}
Note that \textbf{LP}, \textbf{DeepWalk}, \textbf{SemiEmb}, and \textbf{Planetoid} are also baselines in the paper of \textbf{GCN}, we exactly follow their settings in~\cite{kipf2017semi}. In addition, the setting of \textbf{GraphSGAN} is same as the original paper.

\subsubsection{Parameter Settings}
We implement GraphAT and GraphVAT, which train GCN with different versions of graph adversarial training, respectively, with Tensorflow. 
GraphAT has six hyperparameters in total: 1) $D^1$, the size of hidden layer (GCN); 2) $\lambda$, the weight for $L_2$-norm (GCN); 3) dropout ratio (GCN); 4) $\epsilon$, the scale of graph adversarial perturbations (GraphAT); 5) $\beta$, the weight for graph adversarial regularizer (GraphAT); and 6) $K$, the number of sampled neighbors (GraphAT). 
For fair comparison, we set $D^1$, $\lambda$ as the optimal values of standard GCN. But we set dropout ratio as zero in GraphAT for stable training. For the remaining three parameters, $\epsilon$, $\beta$, and $K$, we performed grid-search within the ranges of [0.01, 0.05, 0.1, 0.5, 1], [0.01, 0.05, 0.1, 0.5, 1, 5], [1, 2, 3], respectively.

For GraphVAT, for simplicity, we set the six hyperparameters common to that of GraphAT using the optimal values found for GraphAT. Here we only tune its three additional hyperparameters: 1) $\epsilon'$, the scale of virtual adversarial perturbations; 2) $\alpha$, the weight for virtual adversarial regularizer; and 3) $\xi$, the scale to calculate approximation. 
In particular, we perform grid-search within the ranges of [0.01, 0.05, 0.1, 0.5, 1], [0.001, 0.005, 0.01, 0.05, 0.1, 0.5], [1e-6, 1e-5, 1e-4], respectively. It should be noted that the \textit{Uniform} strategy is adopted to sample neighbor nodes if not other specified.


\subsection{Performance Comparison}
\subsubsection{Model Comparison}
\begin{table}[]
\centering
\caption{Performance of the compared methods on the three datasets \wrt accuracy.}
\vspace{-0.3cm}
\label{tab:perf_comp}
\begin{tabular}{c|cccc}
\hline
Category & Method & Citeseer & Cora & NELL \\ \hline \hline
\multirow{2}{*}{Graph} & LP & 45.3 & 68.0 & 26.5 \\ 
& DeepWalk & 43.2 & 67.2 & 58.1 \\ \hline
\multirow{3}{*}{\begin{tabular}[c]{@{}c@{}}+Node\\ Features\end{tabular}} & SemiEmb & 59.6 & 59.0 & 26.7 \\ 
& Planetoid & 64.7 & 75.7 & 61.9\\ 
& GCN & 69.3 & 81.4 & 61.2 \\ \hline 
\multirow{2}{*}{+Adversarial} & GraphSGAN & 73.1 & \textbf{83.0} & --- \\
& GraphVAT & \textbf{73.7} & 82.6 & \textbf{64.7} \\ \hline
\end{tabular}%
\end{table}
We first investigate how effective is the proposed \textit{graph adversarial training} via comparing the performance of GraphVAT (the extended version of GraphAT) with state-of-the-art node classification methods. Table \ref{tab:perf_comp} shows the classification performance of the compared methods on the three datasets regarding accuracy. The performance of LP, DeepWalk, SemiEmb, and Planetoid are taken from the GCN paper~\cite{kipf2017semi} since we follow its settings exactly. 
From the results, we have the following observations:
\begin{itemize}[leftmargin=*]
	\item GraphVAT significantly outperforms the standard GCN, exhibiting relative improvements of 6.35\%, 1.47\%, and 5.72\% on the Citeseer, Cora, and NELL datasets, respectively. As the only difference between GraphVAT and GCN is applying the proposed graph adversarial training, the improvements are attributed to the proposed training method which would enhance the stabilization and generalization of GCN. Besides, the results validate that GraphVAT is effective in solving the node classification task.
	\item GraphVAT achieves comparable performance as that of GraphSGAN, which is the state-of-the-art method of node classification. It demonstrates the efficacy of the proposed method. However, our method could offer a more feasible solution for two reasons: 1) GraphSGAN is based on the standard generative adversarial network, which explicitly plays a mini-max game between a discriminator and a generator (two different networks). This, inevitably, will lead to doubling of the computation of model training and the labor of parameter tuning. 2) For different node classification applications, GraphSGAN needs to be built from scratch, whereas our GraphVAT is a generic solution that can be seamlessly applied to enhance the existing models of the applications.
	\item GraphVAT and GraphSGAN achieve better results in all the cases as compared to the other baselines. On the Citeseer, Cora, and NELL datasets, the relative improvements are at least 6.35\%, 1.97\%, and 4.52\%, respectively. This indicates the effectiveness of adversarial learning, \ie dynamically playing a mini-max game either implicitly (GraphVAT) and explicitly (GraphSGAN) in the training phase. Moreover, the results are consistent with findings in previous work~\cite{goodfellow2015explaining,yu2018learning,he2018adversarial,miyato2018virtual}.
	\item Among the baselines, 1) the methods that jointly account for the graph structure and node features (in the category of \textit{+Node Features}) outperform LP and DeepWalk that only consider graph structure. This suggests further exploration of how to combine the connection patterns and node features more appropriately. 2) As compared to SemiEmb that is a shallow model, Planetoid and GCN achieves significant improvements (from 8.56\% to 131.8\%) in all cases. The improvement is reasonable and attributed to the strong representation ability of neural networks. As such, methods targeting to enhance the graph neural network models, such as the graph adversarial training, will be meaningful and influential in future.
\end{itemize}

\subsubsection{Performance \wrt Node Degree}
\begin{figure}[htb]
	\centering
	\includegraphics[width=0.35\textwidth]{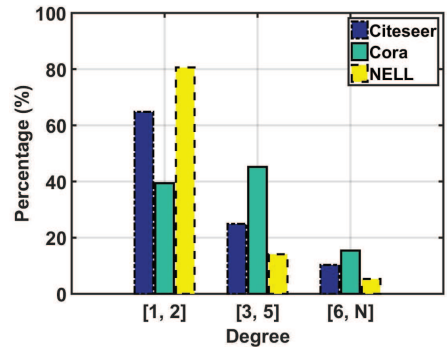} 
	\vspace{-0.2cm}
	\caption{Percentage of nodes with degrees in different groups in the three datasets.}
	\label{fig:group_stat}
\end{figure}

We next study how the graph adversarial training performs on nodes with different densities of connections so as to understand where this regularization technique can be more suitably applied. We empirically split the nodes into three groups according to node degree (\ie the number of neighbors), where node degrees are in ranges of $[1, 2]$, $[3, 5]$, $[6, N]$. Figure~\ref{fig:group_stat} illustrates the distribution of nodes over the three groups. As can be seen, in all the three datasets, a great number of nodes are sparsely connected (with degrees smaller than three), and only about ten percent of the nodes are densely connected with degrees bigger than five. 
\begin{figure*}[]
	\centering
	\mbox{
		\hspace{-0.1in}
		\subfigure[Citeseer]{
			\label{fig:gp_citeseer}
			\includegraphics[width=0.32\textwidth]{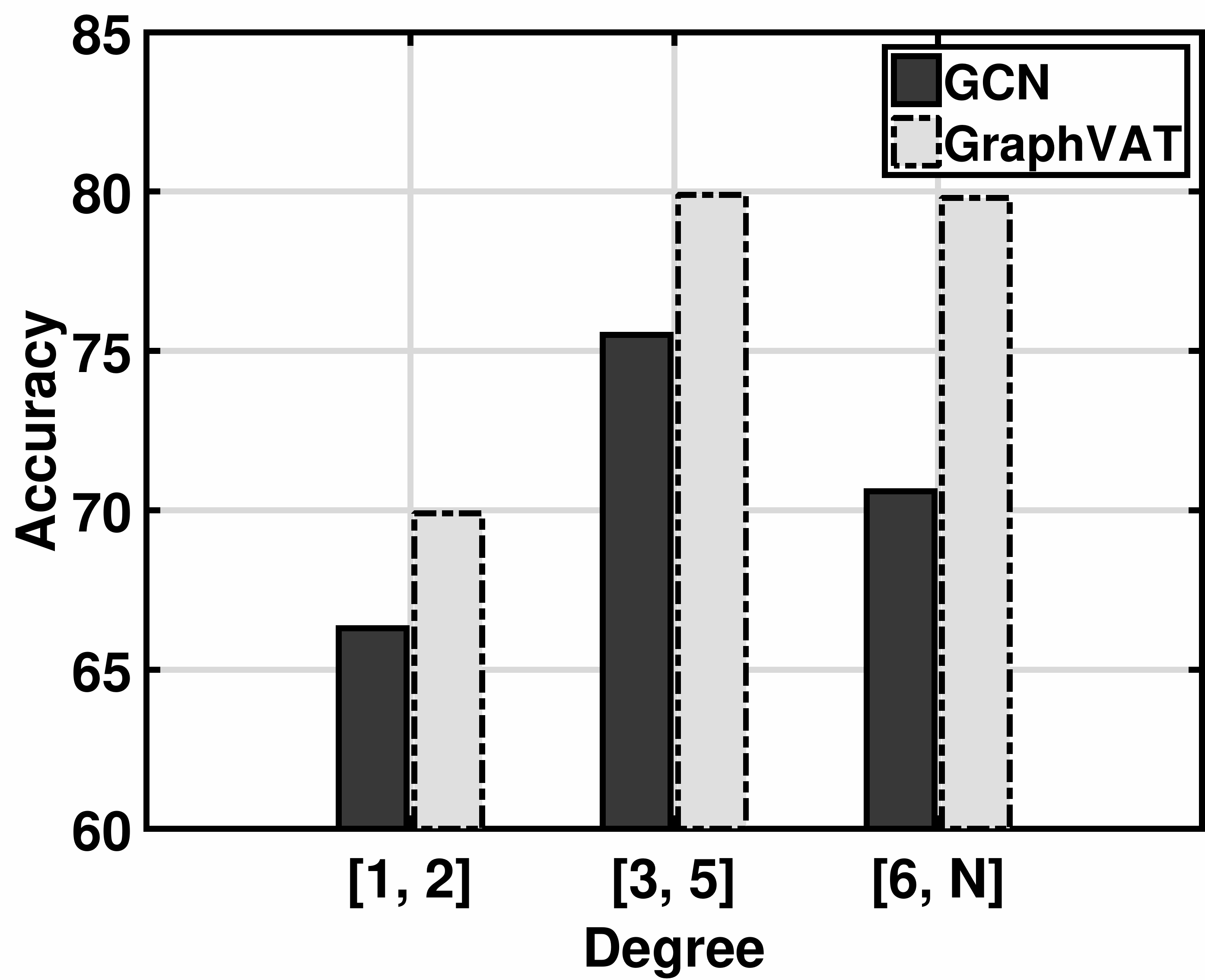}
		}
		\hspace{-0.15in}
		\subfigure[Cora]{
			\label{fig:gp_cora}
			\includegraphics[width=0.32\textwidth]{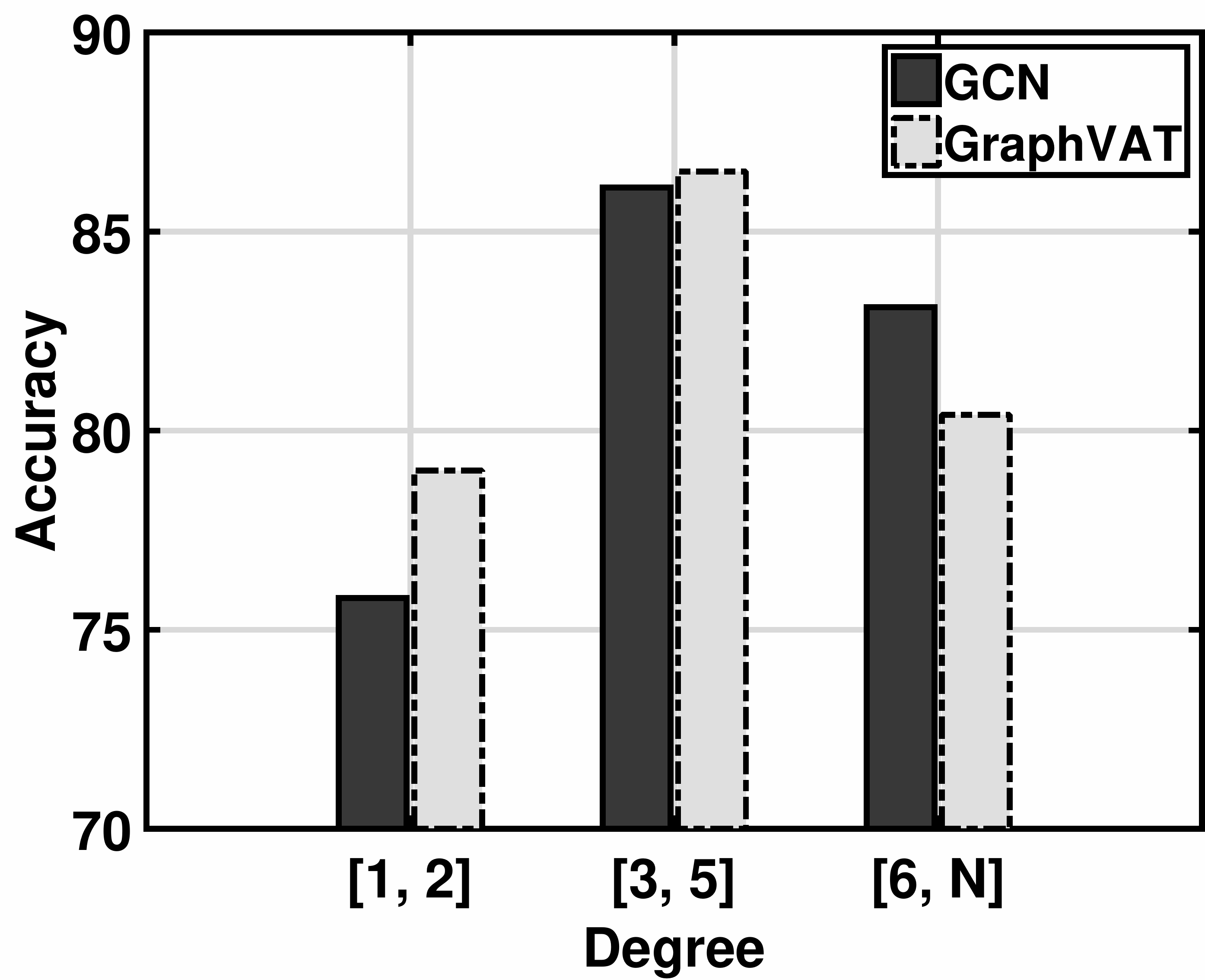}
		}
		\hspace{-0.15in}
		\subfigure[NELL]{
			\label{fig:gp_nell}
			\includegraphics[width=0.32\textwidth]{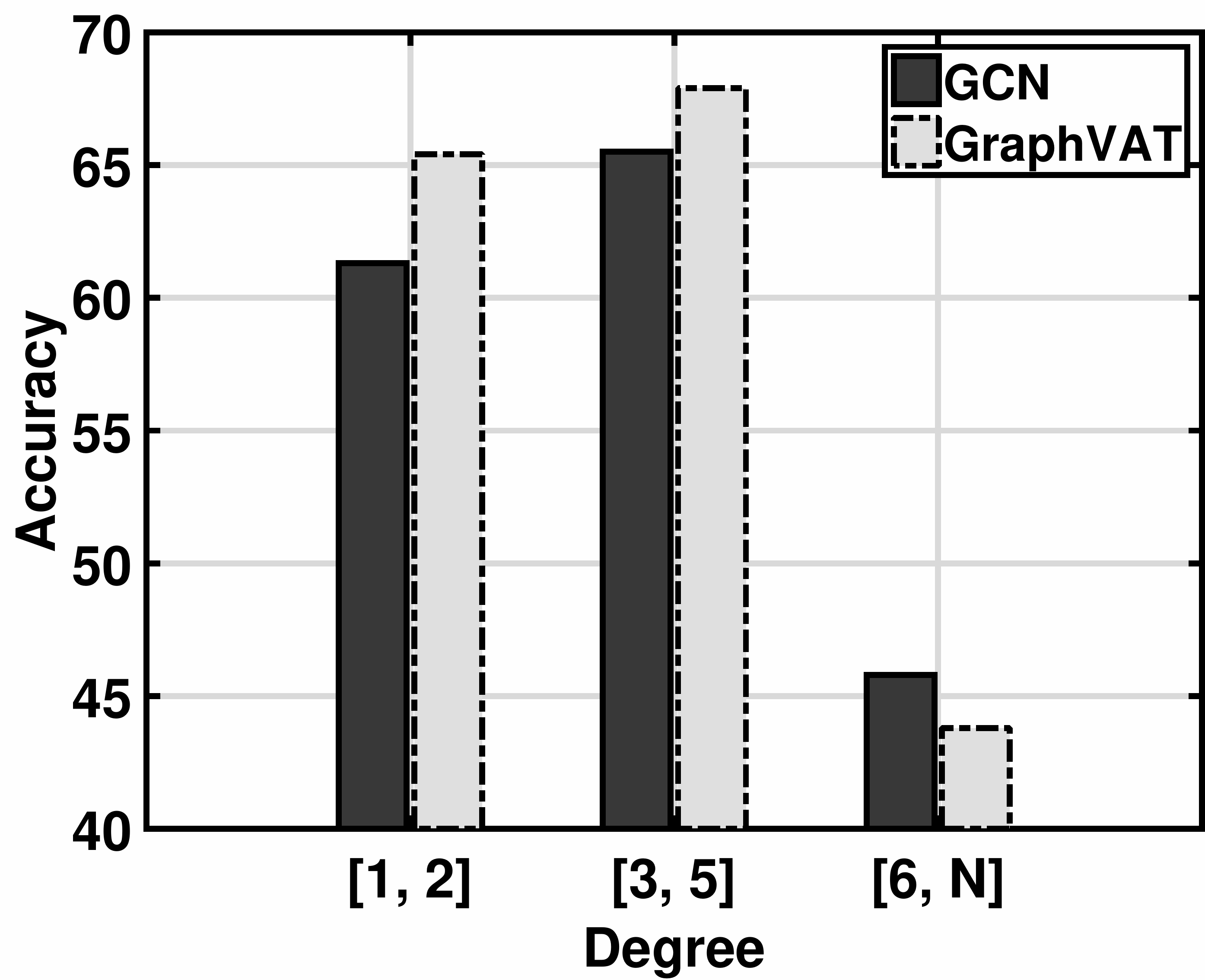}
		}
	}
	\vspace{-0.2cm}
	\caption{Performance of GCN and GraphVAT on nodes with different degrees in Citeseer (a), Cora (b), and NELL (c).
	}
	\label{fig:group_perf}
\end{figure*}
By separately counting the accuracy of GCN and GraphVAT over nodes in different groups, we obtain the group-oriented performance on the three datasets, which is depicted in Figure~\ref{fig:group_perf}. From the results, we observe that:
\begin{itemize}[leftmargin=*]
	\item In all the three datasets, both GCN and GraphVAT achieves the best performance on the group of $[3,5]$. The relatively worse performance on the group of $[1, 2]$ could be attributed to that the nodes in that group are sparsely connected and lacks sufficient signals propagated from the neighbors, which are helpful for the classification~\cite{zhu2003semi,kipf2017semi,gao2018large}. In addition, we postulate the reason for the worse performance over nodes with degrees in $[6, N]$ as such nodes are harder to classify. This is because such nodes typically represent more general entities, such as those having connections to other entities with different types of relations and are thus harder to be accurately classified into a specific category.
	\item In most cases (except the $[6, N]$ group of Cora and NELL), GraphVAT outperforms GCN, which indicates that graph adversarial training would benefit the prediction of nodes with different degrees and is roughly not sensitive to the density of graph. For one of the exceptions (the $[6, N]$ group of NELL), we speculate that the reason is the under-fitting of standard GCN on such nodes (note that the performance of GCN on $[6, N]$ is 27.7\% on average worse than the other two groups), where additional regularization performed by graph adversarial training worsens the under-fitting problem. 
	
	\item GraphVAT significantly and consistently outperforms GCN on the group of $[1, 2]$, with an average improvement of 5.45\%. The result indicates that the graph adversarial training would be more effective on sparse part of the graph. It should be noted that most of the graphs are sparse in real world applications~\cite{cui2018survey}. As such this result further demonstrates the potential of the proposed methods in real world applications.
	
\end{itemize}

\subsubsection{Method Ablation}
\begin{table}[]
\centering
\caption{Effect of graph adversarial regularization and virtual graph adversarial regularization.}
\vspace{-0.2cm}
\label{tab:abla_perf}
\begin{tabular}{c|cccc}
\hline
Category & Method & Citeseer & Cora & NELL \\ \hline \hline
Standard Training & GCN & 69.3 & 81.4 & 61.2 \\ \hline 
\multirow{3}{*}{\begin{tabular}[c]{@{}c@{}}Adversarial\\Training\end{tabular}} & GCN-VAT & 72.4 & 79.3 & 63.3 \\
& GraphAT & 73.4 & 82.5 & 62.3 \\
& GraphVAT & \textbf{73.7} & \textbf{82.6} & \textbf{64.7} \\ \hline
\end{tabular}%
\end{table}
Recall that we design two versions of graph adversarial training: 1) basic GraphAT (Equation~\ref{eqn:objgat}) and 2) incorporating virtual adversarial training (Equation~\ref{eqn:objgatv}). To evaluate the contribution of these two types of regularizations, we compare the performance of the following solutions built upon GCN:
\begin{itemize}[leftmargin=*]
	\item \textbf{GCN}: It learns the parameters of GCN with standard training, \ie it optimizes Equation~\ref{eqn:objgcn}.
	\item \textbf{GCN-VAT}: Virtual adversarial training, which performs perturbations by considering node features only, is employed to train GCN, \ie optimizing Equation~\ref{eqn:objgatv} with $\beta=0$.
	\item \textbf{GraphAT}: It trains GCN by the basic GraphAT, of which the perturbations only focus on graph structure. That is optimizing Equation~\ref{eqn:objgatv} with $\alpha=0$.
	\item \textbf{GraphVAT}: It accounts for both the virtual and graph adversarial regularizations during the training of GCN.
\end{itemize}

Table~\ref{tab:abla_perf} shows the performance of the compared methods on the three datasets \wrt accuracy. As can be seen:
\begin{itemize}[leftmargin=*]
	\item In most of the cases, GCN performs worse than the other approaches, which indicates that adversarial training could enhance the node classification model as compared to the standard training. That is, by intentionally and dynamically generating perturbations and optimizing additional regularizers, the trained model could by more accurate.

	\item GraphVAT achieves the best performance in all cases, \ie GraphVAT outperforms both GCN-VAT and GraphAT. It shows that perturbations targeting at both the individual nodes (virtual adversarial perturbations) and neighbor nodes (graph adversarial perturbations) benefit the training of graph neural network model. Moreover, it suggests that it is beneficial to jointly consider both the node features and graph structure in adversarial training of graph neural networks.
	
	\item Compared to GCN-VAT, GraphAT achieves improvements of 1.38\% and 4.04\% on the Citeseer and Cora datasets, which signifies the benefit of accounting for the graph structure in adversarial training of graph neural networks. However, on the NELL dataset, the performance of GraphAT is 1.58\% worse than GCN-VAT. We speculate that the decrease is mainly because NELL is a bipartite graph where the neighbors of an entity node are all relation nodes without bag-of-words descriptions and labels. Therefore, as compared to standard graph with homogeneous nodes, the generated graph adversarial perturbations according to the predictions of connected relation nodes are less effective. It should be noted that, by resisting such perturbations, GraphAT still implicitly encourages smooth predictions of entity nodes connected by the same relation node, which could be the reason why GraphAT outperforms standard GCN on NELL.
\end{itemize}
\subsubsection{Effect of Sampling Strategies}
\begin{table}[]
\centering
\caption{Performance comparison of GraphAT with different neighbor sampling strategies during adversarial example generation. RI denotes the relative improvement over the Uniform strategy.}
\vspace{-0.2cm}
\label{tab:samp_perf}
\begin{tabular}{c|cccc}
\hline
Sampling Strategy & Citeseer & Cora & NELL & RI \\ \hline \hline
Uniform & 73.4 & 82.5 & \textbf{62.3} & - \\ \hline
Degree & 73.0 & 82.9 & 61.8 & -0.9\% \\
Degree-Reverse & \textbf{73.8} & 82.4 & 62.1 & 0.1\% \\
PageRank & 72.6 & \textbf{83.1} & 62.0 & -0.8\% \\ \hline
\end{tabular}%
\end{table}
As mentioned in Section~\ref{ss:time_comp}, different strategies could be adopted to sample neighbor nodes for the generation of graph adversarial perturbations and the calculation of graph adversarial regularizer. Here, we investigate the effect of sampling strategies via comparing the results of GraphAT performing different samplings. 
Note that we select GraphAT rather than GraphVAT for the reason that we focus on investigating properties of the proposed graph adversarial training.

Table~\ref{tab:samp_perf} shows the corresponding performance, from which we can observe that the performance of different sampling strategies are comparable to each other. 
{As compared to \textit{Uniform}, the relative improvement (RI) achieved by the other strategies is within a range of [-0.9\%, 0.1\%]. 
As such, \textit{Uniform} would be a suitable selection since it will not bring any additional computation as compared to the other approaches.}


\subsection{Effect of Hyperparameters}
\begin{figure*}[]
	\centering
	\mbox{
		\hspace{-0.1in}
		\subfigure[Weight of GraphAT regularization]{
			\label{fig:tune_beta}
			\includegraphics[width=0.3\textwidth]{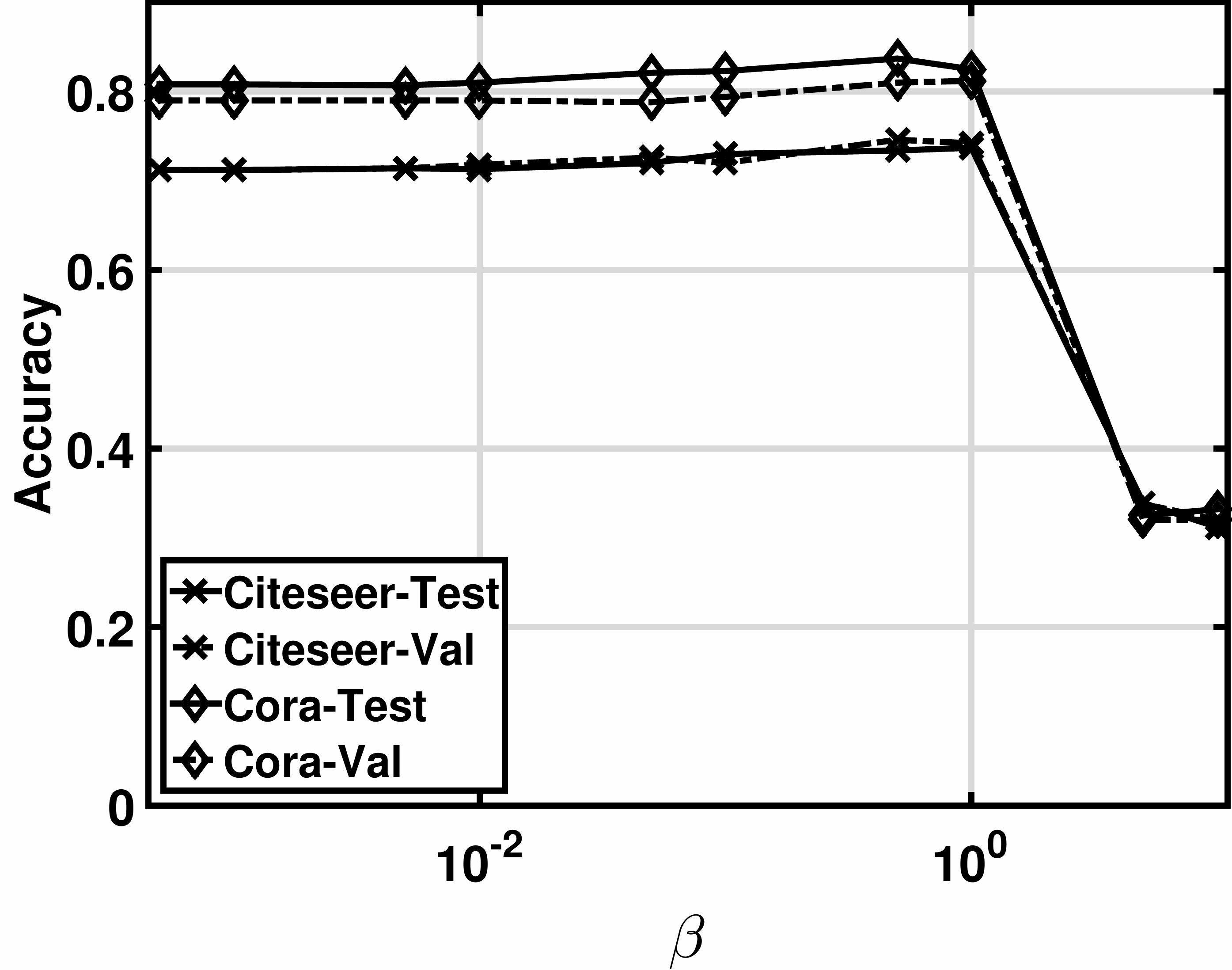}
		}
		\hspace{-0.15in}
		\subfigure[Scale of perturbation]{
			\label{fig:tune_epsilon}
			\includegraphics[width=0.3\textwidth]{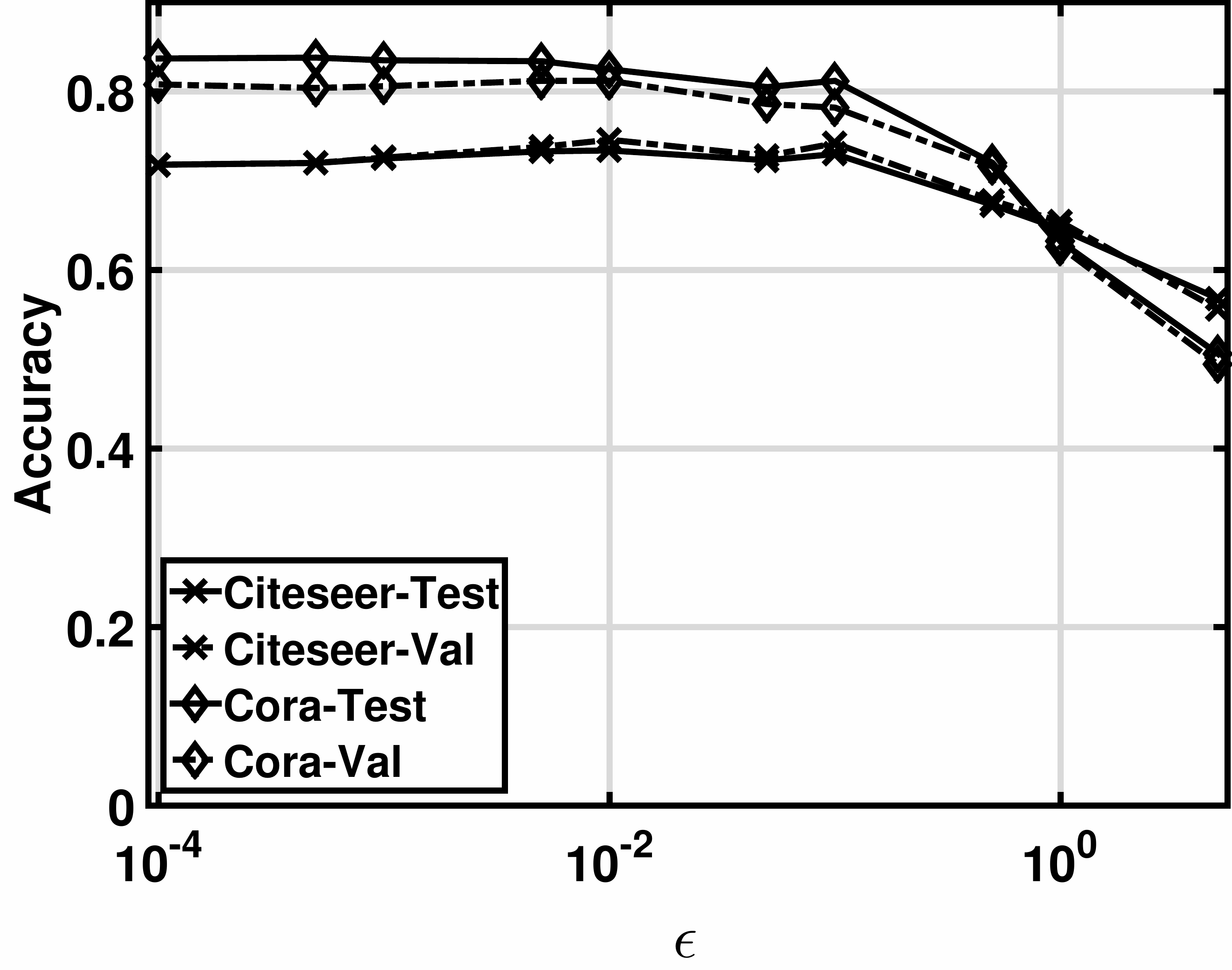}
		}
		\hspace{-0.15in}
		\subfigure[Number of sampled neighbors]{
			\label{fig:tune_k}
			\includegraphics[width=0.3\textwidth]{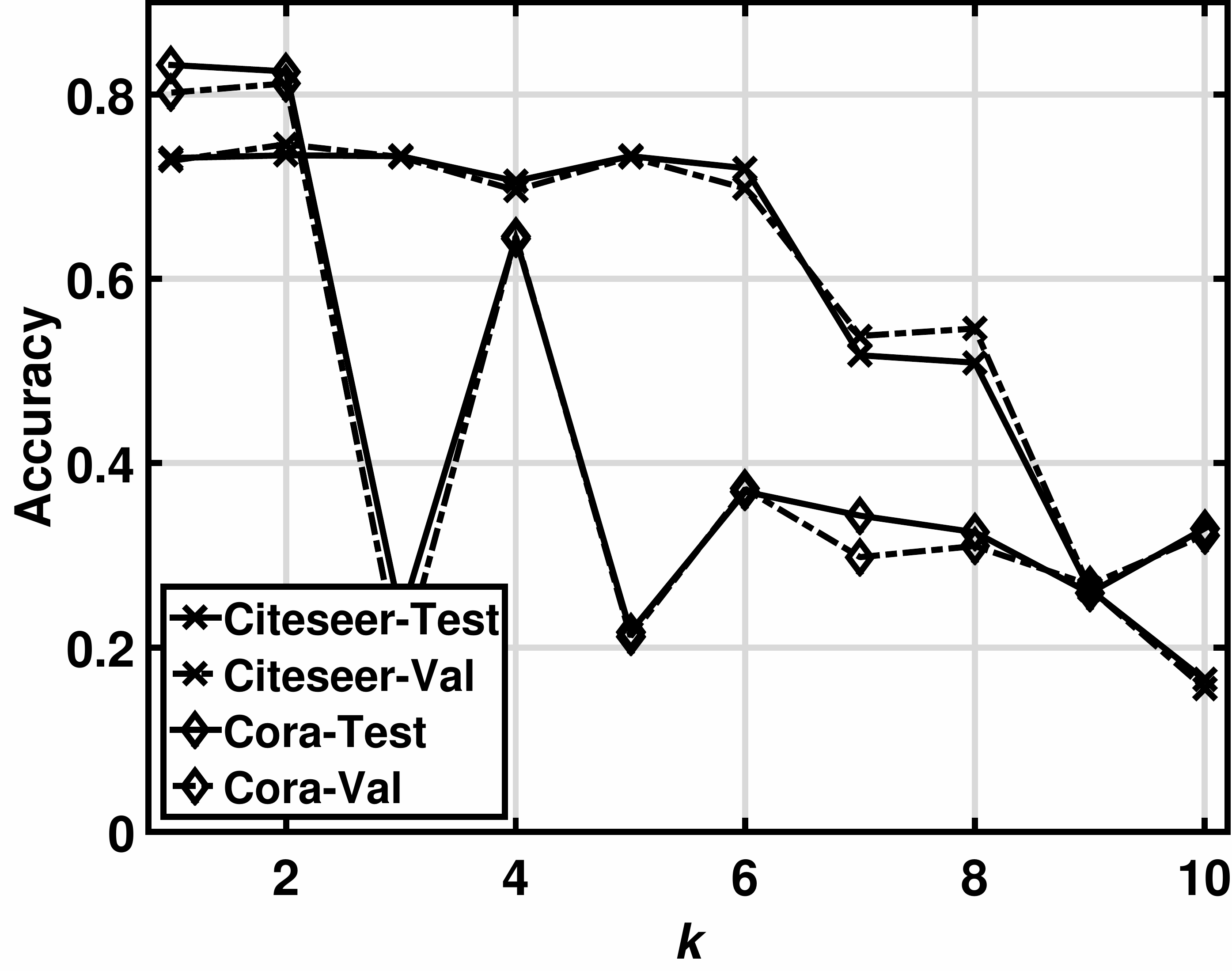}
		}
	}
	\vspace{-0.3cm}
	\caption{Performance of GraphAT with different values of hyperparameters: (a) weight of graph adversarial regularizer ($\beta$), (b) scale of graph adversarial perturbations ($\epsilon$), and (c) number of sampled neighbors ($k$) on the validation and testing of the three datasets (When investigating the effect of a hyperparameter, the other two are set as the optimal values).
	}
	\label{fig:para_tune}
\end{figure*}
\subsubsection{Sensitivity}
We then investigate how the value of hyperparameters effects the performance of the proposed GraphAT. Given a hyperparameter, we evaluate model performance when adjusting its value and fixing the other hyperparameters with optimal values. 
Since focusing on graph adversarial regularization, we mainly study: a) weight of graph adversarial regularizer ($\beta$), b) scale of graph adversarial perturbations ($\epsilon$), and c) number of sampled neighbors ($k$), and use GraphAT to report the performance. It should be noted that, in the following, we focus on the citation graphs and omit results on NELL, which is a bipartite graph rather than a standard graph.

Figure~\ref{fig:para_tune} illustrates the performance of GraphAT on the validation and testing of the three datasets when varying the value of $\beta$, $\epsilon$, and $k$. From the figures, we have the following observations:
\begin{itemize}[leftmargin=*]
	\item Under most cases, the performance of GraphAT changes smoothly near the optimal value of the selected hyperparameter, which indicates that GraphAT is not sensitive to hyperparameters. The only exception is that GraphAT performs significantly worse when $k=3$ and $k=5$ as compared to the performance with other values of $k$. We check the training procedure and observe that both of them are caused by triggering early stopping at the early stage of the training (dozens of epochs), which is occasional and would converge to an expected performance if disable early stopping.
	\item For each parameter, a) GraphAT achieves best performance with $\beta$ is around 0.1, which roughly balances the contribution of the supervised loss and the graph adversarial regularizer (note that the supervised loss decreases fast at the early epochs). Larger value of $\beta$ (stronger regularization) will harm GraphAT since the model could suffer from the underfitting issue. b) GraphAT performs well when $\epsilon$ is in the range of [1e-4, 1e-2], but the performance decreases significantly as increasing $\epsilon$. This result supports the assumption that perturbations have to be in a small scale so that the constructed adversarial examples have similar feature distributions as the real data. c) GraphAT performs best when $k=1$ or $k=2$, which is somehow coherent with the result in Figure~\ref{fig:group_perf} that graph adversarial training are more effective to nodes with degree in [1, 2]. This result is appealing since the computation cost linearly increases as sampling more neighbors.
\end{itemize}

\subsubsection{Tuning $\epsilon$ Only}
\begin{table}[]
	\centering
	\caption{Performance of GraphAT as tuning all hyperparameters (\ie $\beta$, $\epsilon$, and $k$) and tuning $\epsilon$ with fixed $\beta = 1.0$ and $k=1$.}
	\vspace{-0.2cm}
	\label{tab:epsilon_perf}
	\begin{tabular}{c|cc}
		\hline
		Hyperparameter & Citeseer & Cora \\ \hline \hline
		$\{\beta,~\epsilon,~k\}$ & 73.4 & \textbf{82.5} \\
		$\{\epsilon\}$ & \textbf{73.6} & \textbf{82.5} \\ \hline
	\end{tabular}%
\end{table}
Considering that the number of candidate combinations exponentially increases with the number of hyperparameters, we explore whether comparable performance could be achieved when tune one hyperparameter alone and fix the others with empirical values. It should be noted that previous work~\cite{miyato2018virtual} has shown that tuning $\epsilon'$ alone could suffice for achieving satisfactory performance of VAT. Similarly, we tune $\epsilon$ with $\beta = 1$ and $k=1$ and summarize the performance of GraphAT in Table~\ref{tab:epsilon_perf}. As can be seen, on the citation graphs, tuning $\epsilon$ alone achieves satisfactory performance. 
As such, the overhead of additional hyperparameters of the proposed GraphAT could be ignored.

\subsection{Impact of Graph Adversarial Training}
\subsubsection{Training Process}
\begin{figure*}[]
	\centering
	\mbox{
		\hspace{-0.1in}
		\subfigure[Validation (Citeseer)]{
			\label{fig:train_citeseer_val}
			\includegraphics[width=0.26\textwidth]{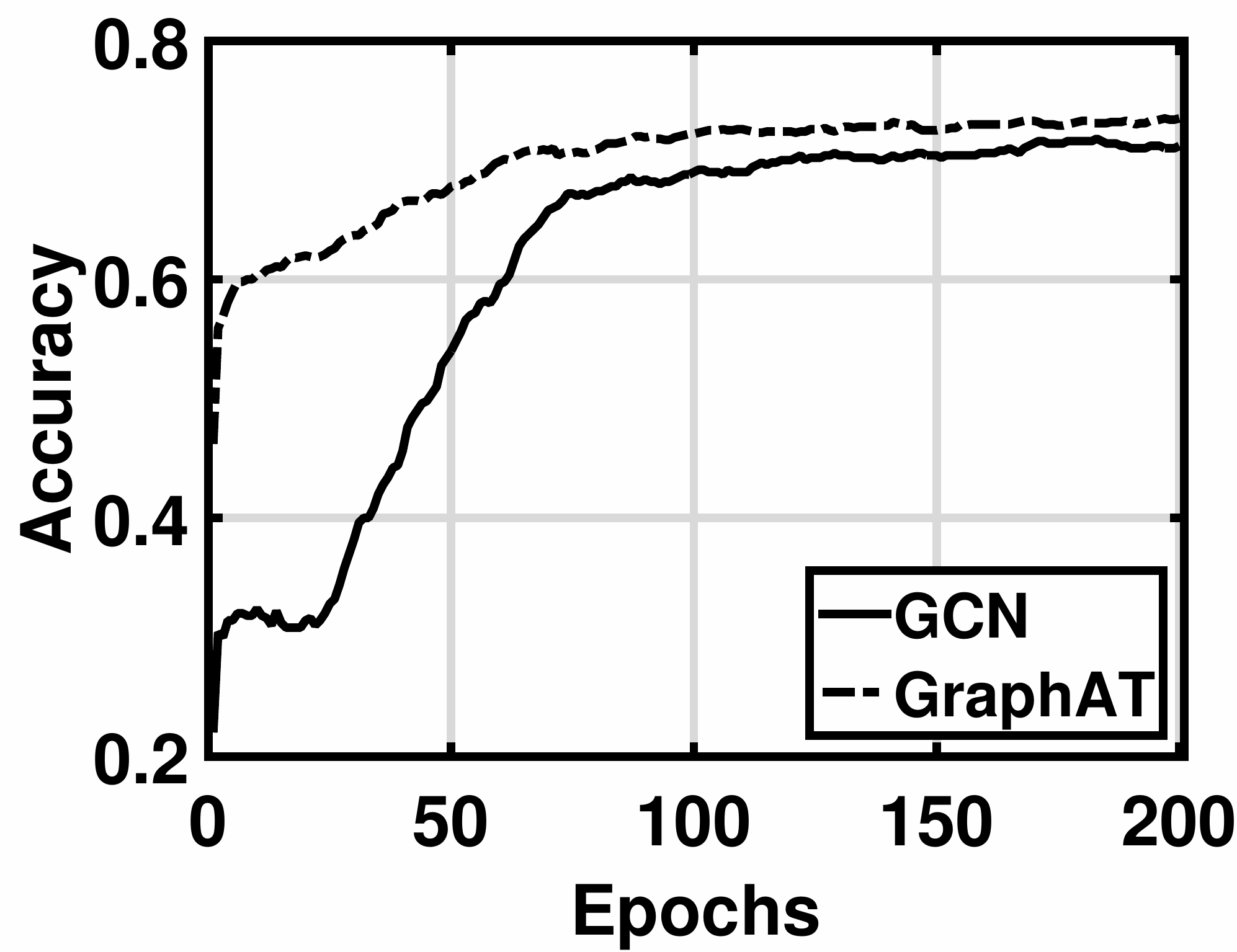}
		}
		\hspace{-0.15in}
		\subfigure[Testing (Citeseer)]{
			\label{fig:train_citeseer_tes}
			\includegraphics[width=0.26\textwidth]{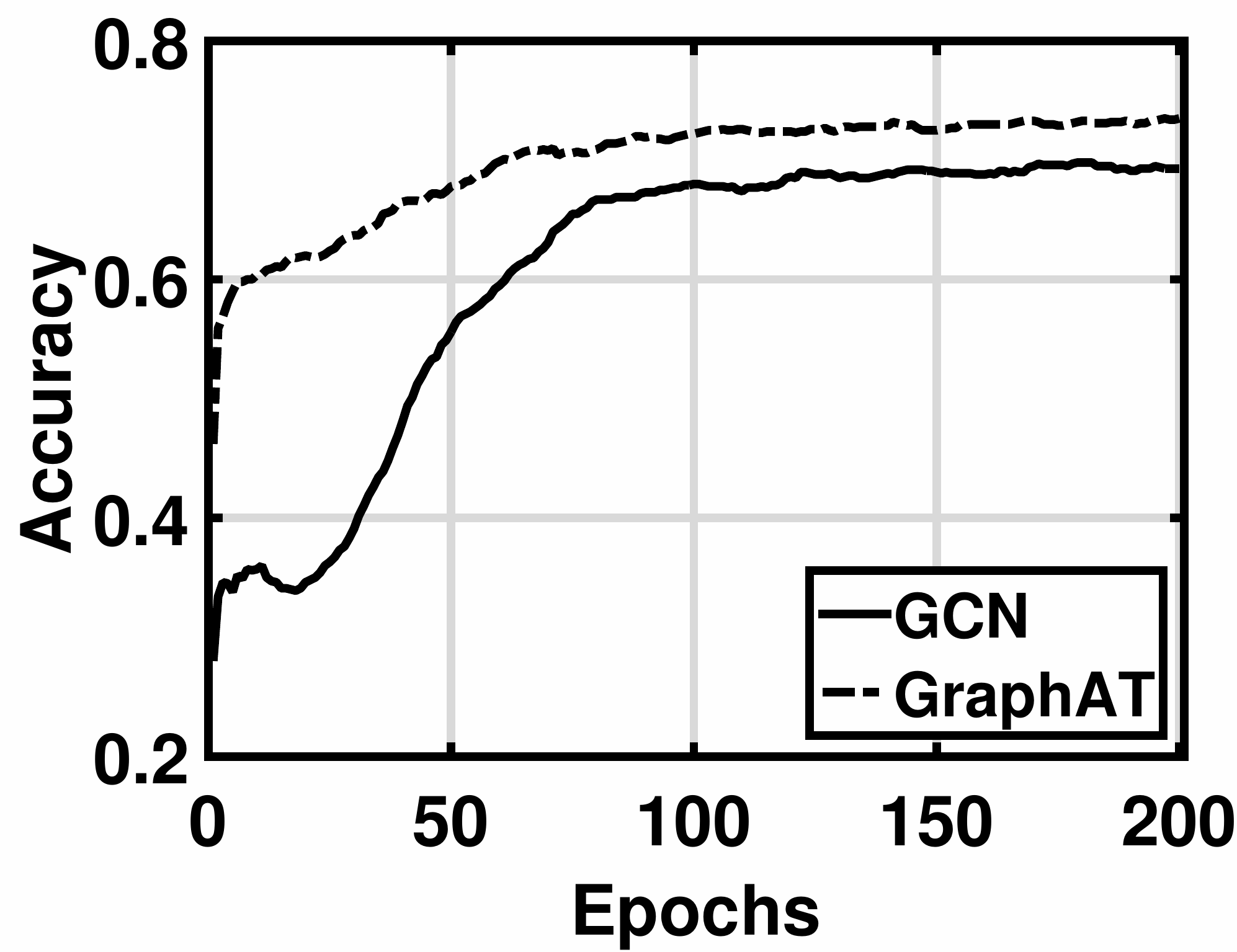}
		}
		\hspace{-0.15in}
		\subfigure[Validation (Cora)]{
			\label{fig:train_cora_val}
			\includegraphics[width=0.26\textwidth]{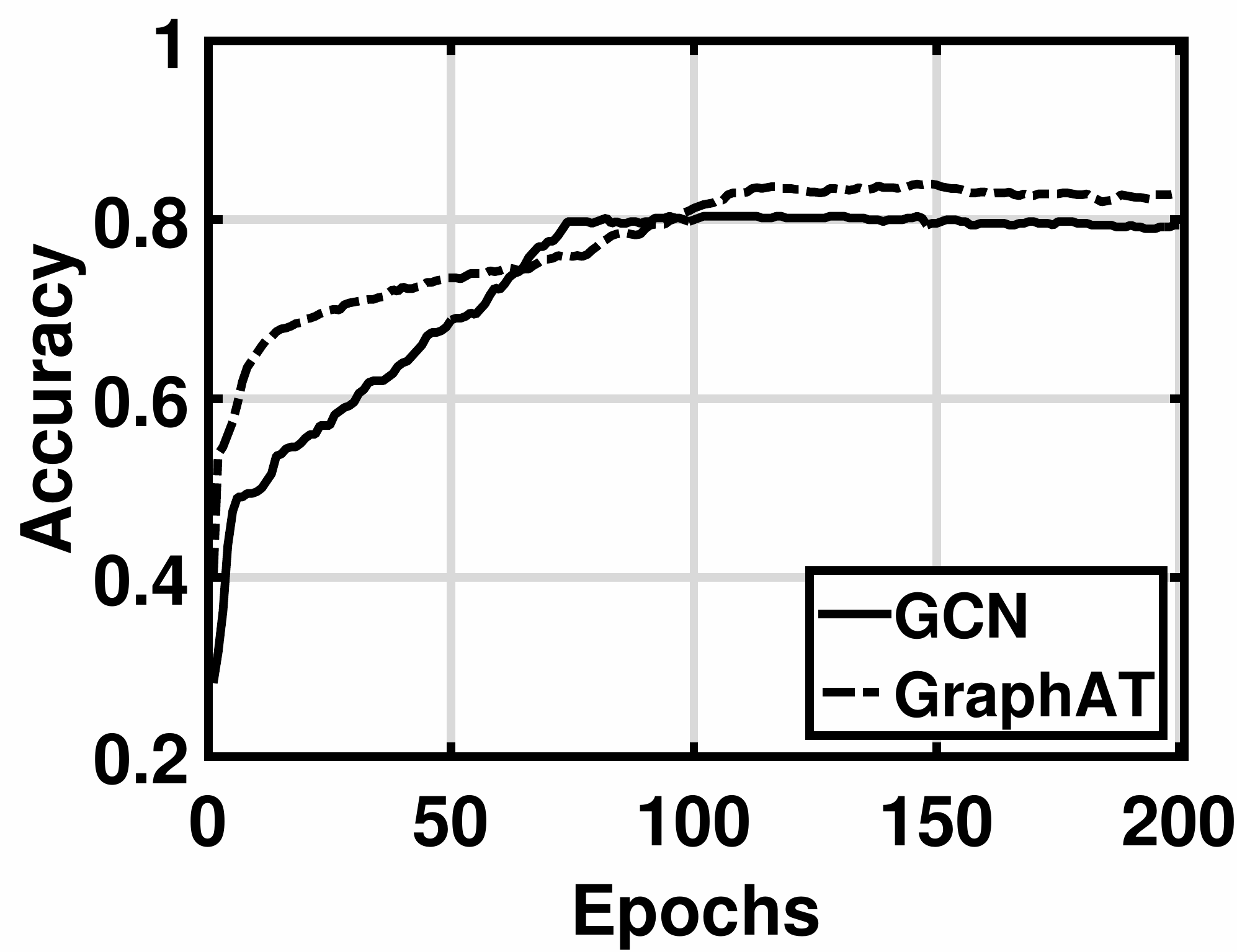}
		}
		\hspace{-0.15in}
		\subfigure[Testing (Cora)]{
			\label{fig:train_cora_tes}
			\includegraphics[width=0.26\textwidth]{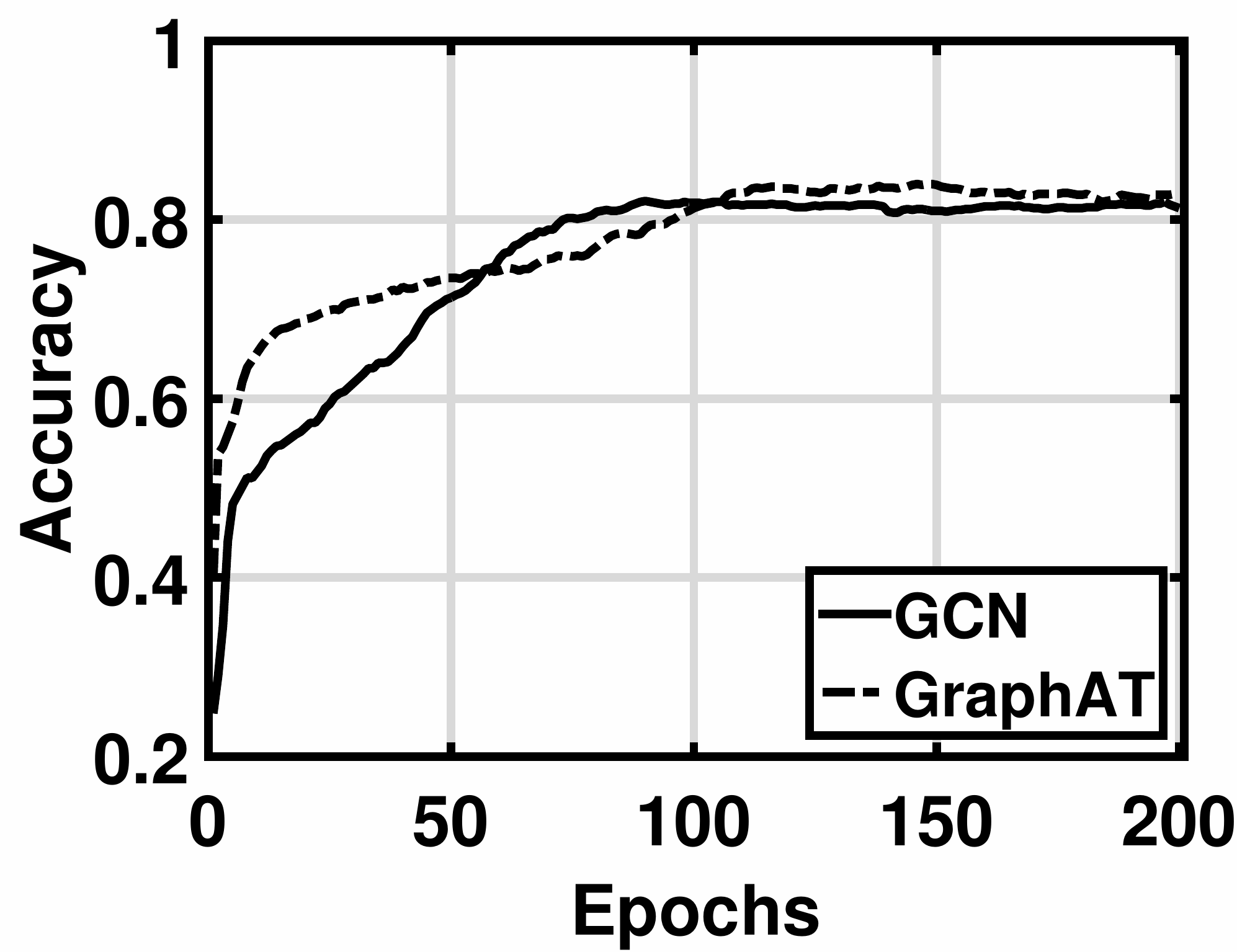}
		}
	}
	\vspace{-0.2cm}
	\caption{Training curves of the GCN and GraphAT on the validation and testing of Citeseer and Cora.
	}
	\label{fig:train_curve}
\end{figure*}
We next study the effect of GraphAT on the training process of GCN. Specifically, we observe the performance of GCN and GraphAT on the validation and testing of Citeseer and Cora after every training epoch, which is depicted in Figure~\ref{fig:train_curve}. 
As can be seen, 1) On the two datasets, the performance of GCN and GraphAT becomes stable after 100 epochs, which indicates that graph adversarial training will not affect the convergence speed of GCN. 2) It is interesting to see that the performance of GraphAT increases faster than standard GCN during the initial several epochs. Note that the supervised loss is typically much larger (about 1e5 times) than the value of graph adversarial regularizer at the initial epochs. This is because all nodes are assigned predictions close to random at the beginning which leads to tiny divergence between connected nodes. As such, the acceleration of performance increase at the initial epochs is believed to be the effect of data augmentation (additional adversarial examples) rather then a better regularization.

\begin{table}[]
	\centering
	\vspace{-0.2cm}
	\caption{The impact of adding graph adversarial perturbations to GCN and GraphAT. The number shows the relative decrease of testing accuracy.}
	\vspace{-0.3cm}
	\label{tab:adv_perf}
	\begin{tabular}{c|cc}
		\hline
		Method & Citeseer & Cora \\ \hline \hline
		GCN & -21.1\% & -6.6\% \\
		GraphAT & -4.1\% & -1.6\% \\ \hline
	\end{tabular}%
\end{table}

\subsubsection{Robustness against Adversarial Perturbations}
Recall that our target is to enhance the robustness of graph neural networks. Table~\ref{tab:adv_perf} shows relative performance decrease of GCN and GraphAT on adversarial examples as compared to clean examples. As can be seen, by training GCN with graph adversarial training, the model becomes less sensitive to graph adversarial perturbations. Graph adversarial perturbations in the scale of 0.01 (\ie $\epsilon=0.01$) decreases accuracy of GCN by 13.9\% on average, while the number is only 2.9\% for GraphAT. It validates that the graph adversarial training technique could enhance the robustness of a GCN model.

\subsubsection{Effect of Graph Adversarial Training on Divergence of Neighbor Nodes}
\begin{table}[]
    \centering
	\caption{Average Kullback-Leibler divergence between connected node pairs calculated from predictions of GCN and GraphAT (small value indicates close predictions).}
	\vspace{-0.2cm}
	\label{tab:kld}
		\begin{tabular}{c|cccc}
			\hline
			Method & \multicolumn{2}{c}{Citeseer} & \multicolumn{2}{c}{Cora} \\ 
			& Test         & All         & Test       & All        \\ \hline \hline
			GCN & 0.132 & 0.137 & 0.345 & 0.333 \\
			GraphAT & 0.127 & 0.130 & 0.308 & 0.299 \\ \hline
		\end{tabular}%
\end{table}
We retrospect that the intuition of graph adversarial regularizer is to encourage connected nodes to be predicted similarly. Table~\ref{tab:kld} shows the effect of applying graph adversarial training to train GCN, from which we can see that graph adversarial training reduces the divergence between connected nodes as expected. These results show that the predictions of GraphAT are more smooth over the graph structure, which indicates the stronger generalization ability and robustness of the trained model.
\section{Conclusion}
In this work, we proposed a new learning method, named \textit{graph adversarial training}, which additional accounts for relation between examples as compared to standard adversarial training. By iteratively generating adversarial examples attacking the graph smoothness constraint and learning over adversarial examples, the proposed method encourages the smoothness of predictions over the given graph, a property indicating good generalization of the model. As can be seen as a dynamic regularization technique, our method is generic to be applied to train most graph neural network models. We trained one well-established model, GCN, with the proposed method to solve the node classification task. By conducting experiments on three benchmark datasets, we demonstrated that training GCN with our method is remarkably effective, achieving an average improvement of 4.51\%. Moreover, it also beats GCN trained with VAT, indicating the necessity of performing AT with graph structure considered

In future, we will explore we are interested to explore the effectiveness of GAD on more graph neural network models~\cite{hamilton2017inductive,velickovic2018graph,ying2018hierarchical}. Moreover, we are interested to investigate the effect of GAD on other graph-based learning tasks such as link prediction and community detection. As focusing on graph-based learning with only one graph in this paper, one potential future work is to investigate the effectiveness of graph adversarial training for graph-based learning methods simultaneously handling multiple graphs. In addition, we are interested in testing the performance of graph adversarial training on graphs with specifical structures, for instance, hyper-graphs and heterogeneous information graphs. Moreover, we would like to incorporate techniques like robust optimization~\cite{madry2017towards} and adversarial dropout~\cite{park2018adversarial} into the proposed method to further enhance its ability of stabilizing graph neural network models.

\section*{Acknowledgments}
This research is part of NExT++ reserach, supported by the National Research Foundation Singapore under its AI Singapore Programme (AISG-100E-2018-012), Prime Minister's Office under its IRC@SG Funding Initiative, and National Natural Science Foundation of China (61972372). 
We thank the anonymous reviewers and the associated editor for their reviewing efforts.

\bibliographystyle{IEEEtran}
\bibliography{main}
\vspace{-34pt}
\begin{IEEEbiography}
[{\includegraphics[width=1.0in,height=1.4in,clip,keepaspectratio]{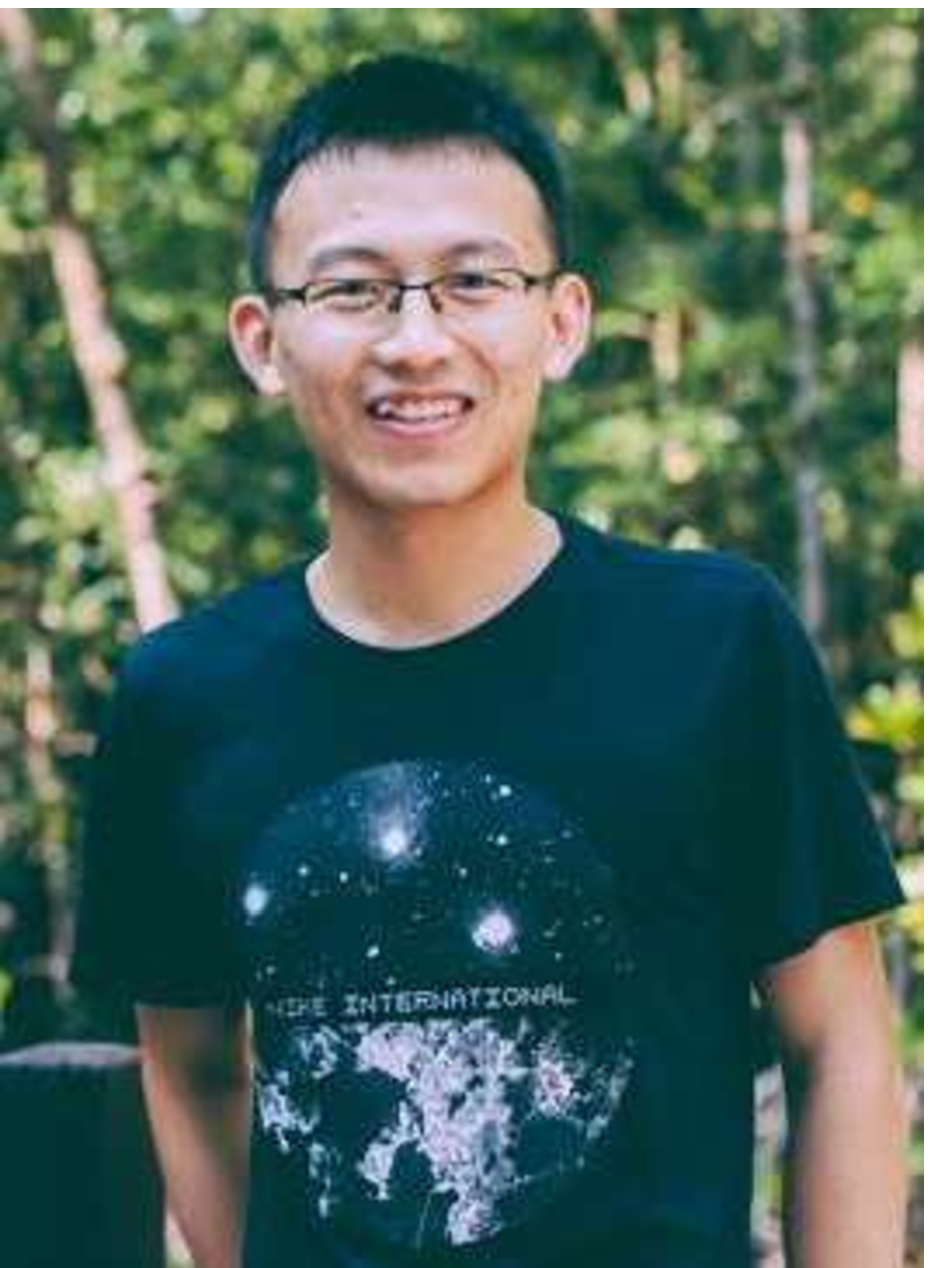}}]{Fuli Feng} is a Ph.D. student in the School of Computing, National University of Singapore. He received the B.E. degree in School of Computer Science and Engineering from Baihang University, Beijing, in 2015. His research interests include information retrieval, data mining, and multi-media processing. He has over 10 publications appeared in several top conferences such as SIGIR, WWW, and MM. His work on Bayesian Personalized Ranking has received the Best Poster Award of WWW 2018. Moreover, he has been served as the PC member and external reviewer for several top conferences including SIGIR, ACL, KDD, IJCAI, AAAI, WSDM etc.
\end{IEEEbiography}

\vspace{-35pt}
\begin{IEEEbiography}
[{\includegraphics[width=1.0in,height=1.4in,clip,keepaspectratio]{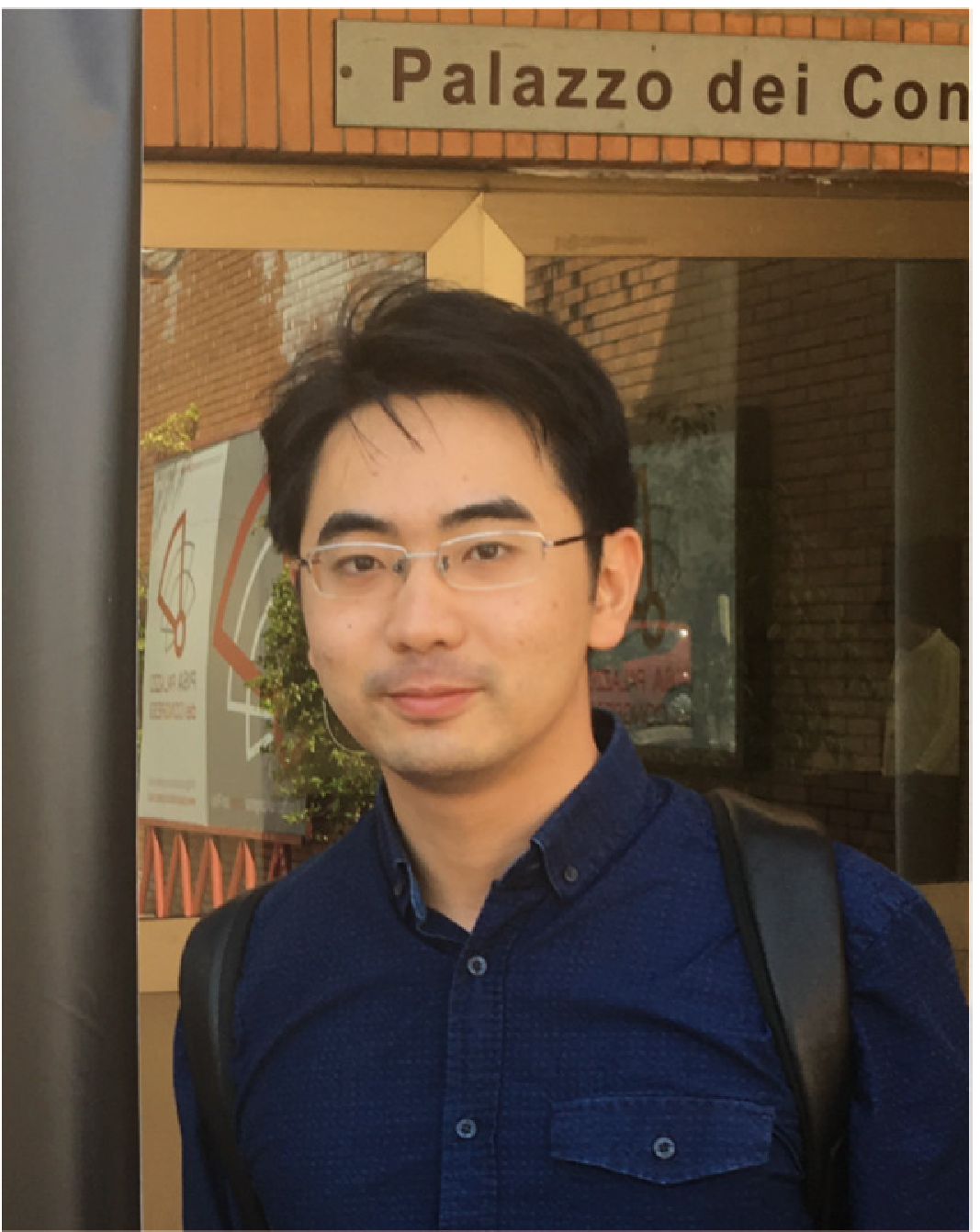}}]{Xiangnan He} is currently a research fellow with School of Computing, National University of Singapore (NUS). He received his Ph.D. in Computer Science from NUS. His research interests span recommender system, information retrieval, natural language processing and multi-media. His work on recommender system has received the Best Paper Award Honorable Mention in WWW 2018 and SIGIR 2016. Moreover, he has served as the PC member for top-tier conferences including SIGIR, WWW, MM, KDD, WSDM, CIKM, AAAI, and ACL, and the invited reviewer for prestigious journals including TKDE, TOIS, TKDD, TMM, and WWWJ. \end{IEEEbiography}
\vspace{-30pt}

\begin{IEEEbiography}
[{\includegraphics[width=1.0in,height=1.4in,clip,keepaspectratio]{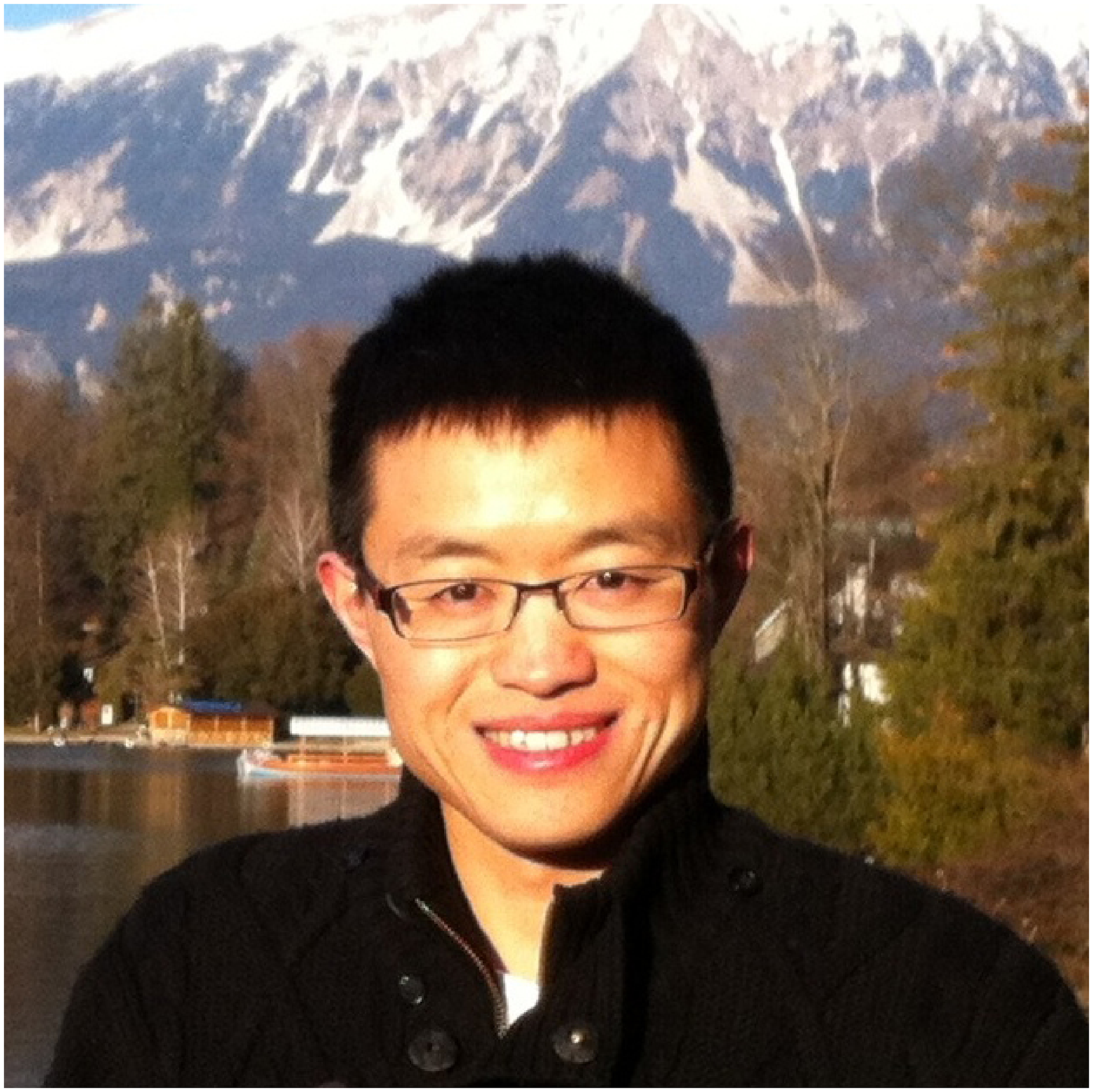}}]{Jie Tang} is an associate professor with the Department of Computer Science and Technology, Tsinghua University. His main research interests include data mining algorithms and social network theories. He has been a visiting scholar with Cornell University, Chinese University of Hong Kong, Hong Kong University of Science and Technology, and Leuven University. He has published more then 100 research papers in major international journals and conferences including: KDD, IJCAI, AAAI, ICML, WWW, SIGIR, SIGMOD, ACL, Machine Learning Journal, TKDD, and TKDE.
\end{IEEEbiography}
\begin{IEEEbiography}[{\includegraphics[width=1.0in,height=1.4in,clip,keepaspectratio]{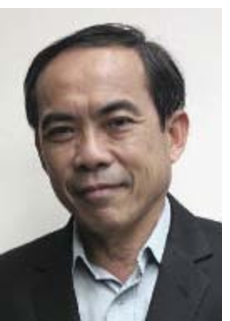}}]{Tat-Seng Chua}
Tat-Seng Chua is the KITHCT (short for Kwan Im Thong Hood Cho Temple) Chair Professor at the School of Computing, National University of Singapore. He was the Acting and Founding Dean of the School from 1998-2000. Dr Chuas
main research interest is in multimedia information
retrieval and social media analytics. In
particular, his research focuses on the extraction,
retrieval and question-answering (QA) of
text and rich media arising from the Web and
multiple social networks. He is the co-Director of
NExT, a joint Center between NUS and Tsinghua
University to develop technologies for live social media search. Dr Chua
is the 2015 winner of the prestigious ACM SIGMM award for Outstanding
Technical Contributions to Multimedia Computing, Communications and
Applications. He is the Chair of steering committee of ACM International
Conference on Multimedia Retrieval (ICMR) and Multimedia Modeling
(MMM) conference series. Dr Chua is also the General Co-Chair of
ACM Multimedia 2005, ACM CIVR (now ACM ICMR) 2005, ACM SIGIR
2008, and ACMWeb Science 2015. He serves in the editorial boards of
four international journals. Dr. Chua is the co-Founder of two technology
startup companies in Singapore. He holds a PhD from the University of
Leeds, UK.
\end{IEEEbiography}

\end{document}